\documentclass[runningheads]{llncs}
\usepackage{graphicx}
\usepackage{hyperref}

\begin{document}
\title{Multi Agent Framework for Collective Intelligence Research}
%
%
\author{
Alexandru Dochian\inst{1}\orcidID{2776000}
}
\authorrunning{Dochian et al.}
%

\institute{
Vrije Universiteit Amsterdam, Amsterdam, The Netherlands
}
\maketitle              

{\centering 
\noindent
\href{https://github.com/alexandru-dochian/multi_agent_framework}{Source code} \\
}

\begin{abstract}
This paper presents a scalable decentralized multi agent framework that facilitates the exchange of information between computing units through computer networks.
The architectural boundaries imposed by the tool make it suitable for collective intelligence research experiments ranging from agents that exchange hello world messages to virtual drone agents exchanging positions and eventually agents exchanging information via radio with real Crazyflie drones in VU Amsterdam laboratory.
The field modulation theory is implemented to construct synthetic local perception maps for agents, which are constructed based on neighbouring agents positions and neighbouring points of interest dictated by the environment.
By constraining the experimental setup to a 2D environment with discrete actions, constant velocity and parameters tailored to VU Amsterdam laboratory, UAV Crazyflie drones running hill climbing controller followed collision-free trajectories and bridged sim-to-real gap. 

\end{abstract}

\section{Introduction}

In recent years, collective intelligence has gained significant attention in science studies due to its potential to cover complex problems through collaboration and coordination among agents \cite{ferrante2012self,valentini2017best}. Unmanned Aerial Vehicle (UAV or drone) technology is being constantly developed and used in-the-wild to aid with tasks such as monitoring agricultural fields \cite{albani2017monitoring,albani2019field} or forests \cite{zhou2022swarm,brust2015networked}, for search and rescue missions \cite{horyna2023decentralized}, firefighting \cite{madridano2021software,hu2022fault} or recreational activities like drone light technology \cite{nar2022optimal}.

Quadcopters are UAVs that have four rotors, each powered by a motor and propeller. Their popularity arises from their high maneuverability, cost-effectiveness in both construction and maintenance, and compact size. These qualities make quadcopters a fine choice for a variety of tasks, whether in areas with limited access or when exploring large regions through coordinated efforts.

Considering the task of monitoring a forest, a swarm of UAVs will take less time to complete it, proportionally with the number of aggregated agents. A larger UAV swarm enhances the overall system reliability by distributing the risk of failure across multiple units.

This work focuses on the high-level decision-making of drones in a 2D space. With access only to local information, the drones follow high values of the perceived gradient while maintaining collision-free trajectories. Each UAV independently chooses a discrete direction to move at a constant velocity, unaffected by the choices of neighboring drones.

A software tool, specifically a \textbf{multi agent framework}, has been developed to support the processing of UAVs in both simulation and deployment in the VU Amsterdam laboratory. This advancement sets the stage for future implementations that can progressively adopt on-the-edge processing directly on the robots.

Following sections will detail the architecture of the proposed framework, the methodologies employed for trajectory planning and swarm cohesion, and the results of preliminary tests carried out in simulated and real-world laboratory environment. These discussions aim to underline the robustness and scalability of the paper’s approach, setting the stage for future research that could extend these principles to other domains of collective intelligence.

\section{Related Work}

Crazyflie drones with limited sensing capabilities managed in \cite{karaguzel2023collective} to collectively follow the gradient of a scalar field without having explicit gradient sensing capabilities. This collective behaviour emerged from individuals UAVs modulating their speeds based on the social interaction with their nearest neighbours. 

In the truly groundbreaking work of \cite{zhou2022swarm}, a swarm of fully autonomous drones is deployed in an unknown dense forest. Each drone has complete onboard systems for perception, localization, and control. The swarm collectively maps the environment while performing real-time spatio-temporal optimization, managing flight paths, avoiding obstacles, and coordinating with other drones. This research represents a significant step towards enabling drone swarms to operate effectively in complex unknown dynamic real-world environments.

A Deep Reinforcement Learning approach for real-time UAV path planning in dynamic environments is detailed in \cite{yan2020towards}. This method utilizes global shared information to generate 84x84 situational maps (it has been an important source of inspiration for this paper). A batch of 12 consecutive maps is used as input for a DQN network (DDQN and D3QN were also benchmarked), which determine the optimal action from a set of 8 discrete 2D movements, with the UAV operating at a constant velocity throughout the simulation.

In addition to the \textbf{multi agent framework} (section \ref{sec:multi-agent-framework}) developed in this work, other systems have been designed to work with Crazyflie drones. Crazyswarm (\cite{preiss2017crazyswarm}) offers its own Python API, abstracting away the communication details with the physical drones. It has proven to be reliable in controlling up to 49 UAVs with high precision, low latency, and the capability to perform synchronized maneuvers.

\section{Methodology}

This section will provide in-depth information about the framework in the first part, and the main theory used for collision-free trajectory planning of the UAV swarm in the second part.

\subsection{Multi Agent Framework}
\label{sec:multi-agent-framework}

This software tool is a runtime environment for healthy execution of independent computational units or \textbf{AppProcess}(es) which are operating system level Threads or Processes.

Interacting with the framework has been simplified a lot through:

\begin{itemize}
    \item the use of installation script tailored for each supporting target operating system: Ubuntu, macOS and Windows.
    \item use of \textit{json configuration files} which hold all the necessary information for initializing an experiment whilst the actual execution logic is already contained in the framework (in custom implementation of the base abstract classes, as presented in next sections). This design choice ensures that experiments can be replicated and parts of them reused. 
\end{itemize}

\subsubsection{Architecture}

\begin{figure}[ht]
  \centering
  \includegraphics[width=1\textwidth]{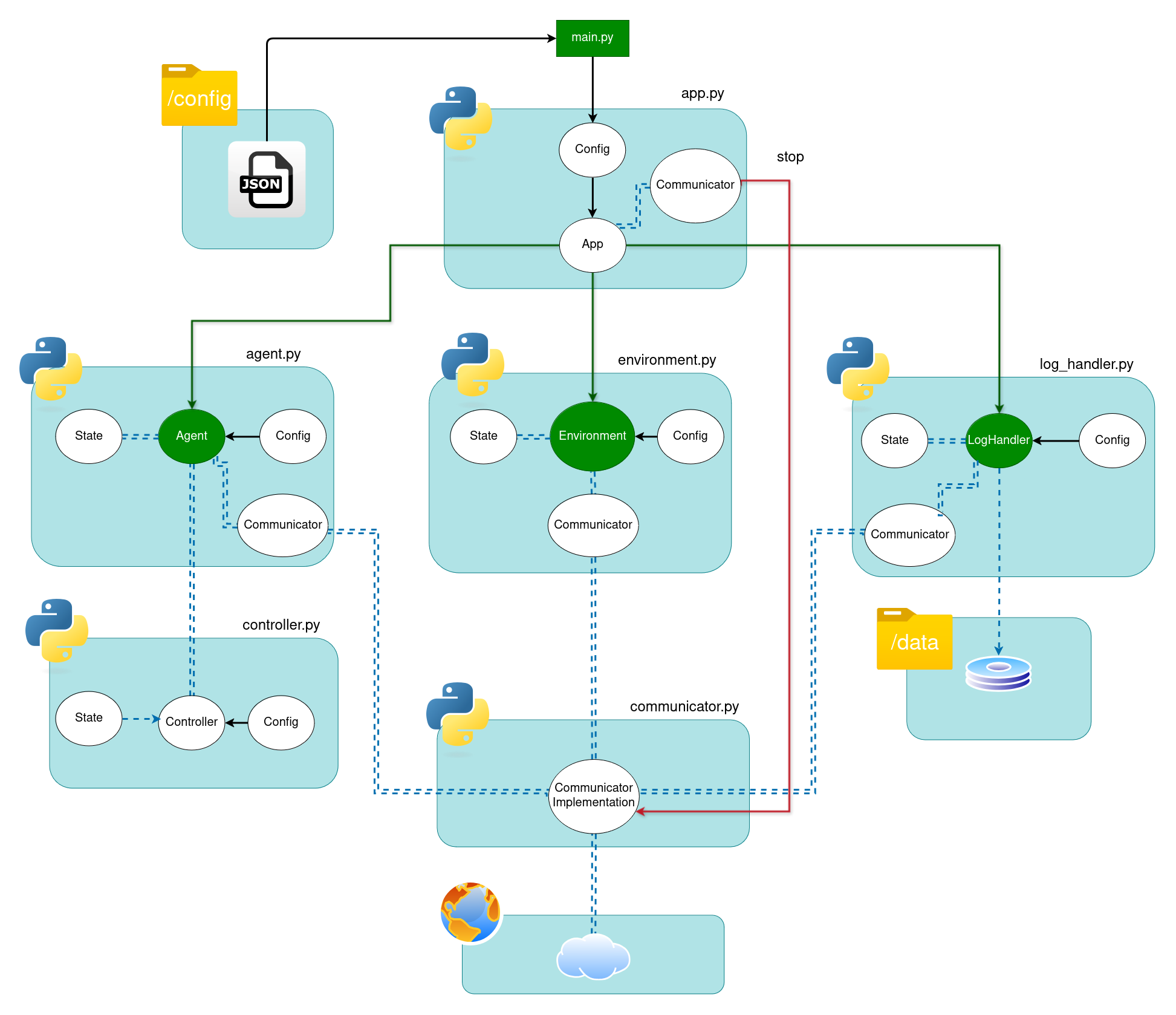}
  \caption{Multi Agent Framework Architecture}
  \label{fig:architecture}
\end{figure}

Fig.~\ref{fig:architecture} presents the common blueprint of any implementation supported by the framework. 

Following legend explanation will maintain its consistency for all the architectural diagrams presented in this paper. 

\begin{itemize}
    \item \textit{Green ellipse} - standalone independent \textbf{AppProcess};
    \item \textit{White ellipse} - object in memory;
    \item \textit{Green line} - spawning a new \textbf{AppProcess};
    \item \textit{Red line} - stop signal is being propagated from the root process to the communicator to be eventually received by all the \textbf{AppProcess}(es);
    \item \textit{Dashed blue line} - directed information exchange;
    \item \textit{Black line} - passing initialization information at instantiation;
    \item \textit{Blue timer icon} - is used to suggest that the logic is executed under a particular delay;
\end{itemize}

The \textbf{config.json} file, that the root process expects, contains all the necessary information for initialization. 
\textbf{AppProcess}(es) will be spawned and objects, along side their typed configuration data transfer objects, will be instantiated in the memory of the corresponding \textbf{AppProcess}.

Any configuration file provides a list of chosen implementation initialization configuration for some of the base classes: \textbf{Agent}(s), \textbf{Controller}(s) \textbf{Environment}(s) and \textbf{LogHandler}(s). It's recommended that any implementation used a typed \textbf{Config} for initialization purposes and a typed \textbf{State} object for modelling its private information in time.

\textbf{Agent}(s) send and receive information in the network through their \textbf{Communicator} and use their underlying \textbf{Controller} for decision making. This separation of concerns is useful when different \textbf{Agent}(s) might use the same \textbf{Controller}(s) or same \textbf{Agent}(s) might use different \textbf{Controller}(s).  

\textbf{Environment}(s) also do send and receive information in the network through their \textbf{Communicator}. Main implementation, which will be presented shortly, places some points of interest in the network with respect to a unique global reference frame for the \textbf{Agent}(s) to then receive and incorporate into their local sensing.

\textbf{LogHandler}(s) recommended responsibility is to retrieve data from the network through the \textbf{Communicator}, store it to disk and, optionally, display it in real-time as in some implementations that will be presented in the sections to come.

The constraints on the particular implementations are rather relaxed, irrespective of base abstract class:
In general they need to provide logic for a \textbf{run()} method or \textbf{predict()} method in the case of \textbf{Controller}(s).

\subsubsection{Hello World experiment}  
\begin{figure}[ht]
  \centering
  \includegraphics[width=1\textwidth]{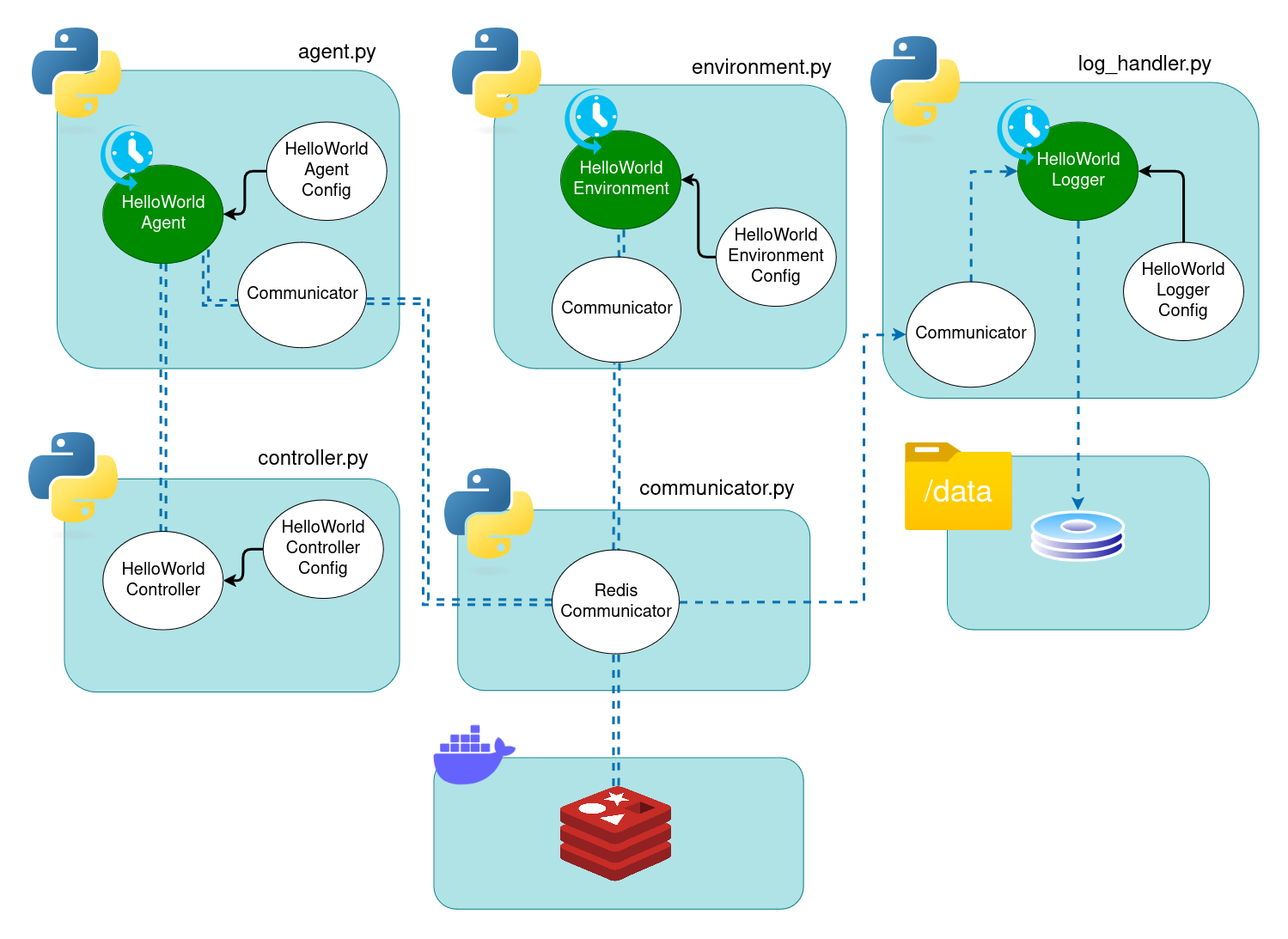}
  \caption{Hello World}
  \label{fig:hello-world}
\end{figure}

makes a good point for the range of possible implementations that the system supports and serves as a compass for anyone wishing to extend the system with a new implementation. Fig.~\ref{fig:hello-world} describes the choices for this experiment.

\textbf{HelloWorldAgent} uses a primitive integer counter state and gets initialized with \textbf{HelloWorldAgentConfig}, which mainly holds the delay at which the logic is being executed. The logic itself is implemented with the use of \textbf{HelloWorldController} (has a primitive integer counter state and gets initialized with \textbf{HelloWorldControllerConfig}) that generates a [\textit{Hello World}] message. This information gets to the agent which appends its own [\textit{Hello World}] message and sends it on the network through its \textbf{Communicator} to be eventually stored in the redis database.
\textbf{HelloWorldEnvironment} (with primitive integer counter state and initialized with \textbf{HelloWorldEnvironmentConfig}) loops at its own configured delay and on each iteration it reads from the network all the \textbf{Agent}(s) messages, appends its own message and writes it back on the network.

\textbf{HelloWorldLogger} (stateless and initialized with \textbf{HelloWorldLoggerConfig}) loops at its own configured delay and on each iteration it reads from the network the message placed by \textbf{HelloWorldEnvironment}, prints it to console and writes it to disk (example of a message retrieved by \textbf{HelloWorldLogger}:  \textit{[[HelloWorldController says hello 1] | [HelloWorldAgent says hello 1]] | [HelloWorldEnvironment says hello 2]} ).

\subsubsection{Default experiment} 
\begin{figure}[ht]
  \centering
  \includegraphics[width=1\textwidth]{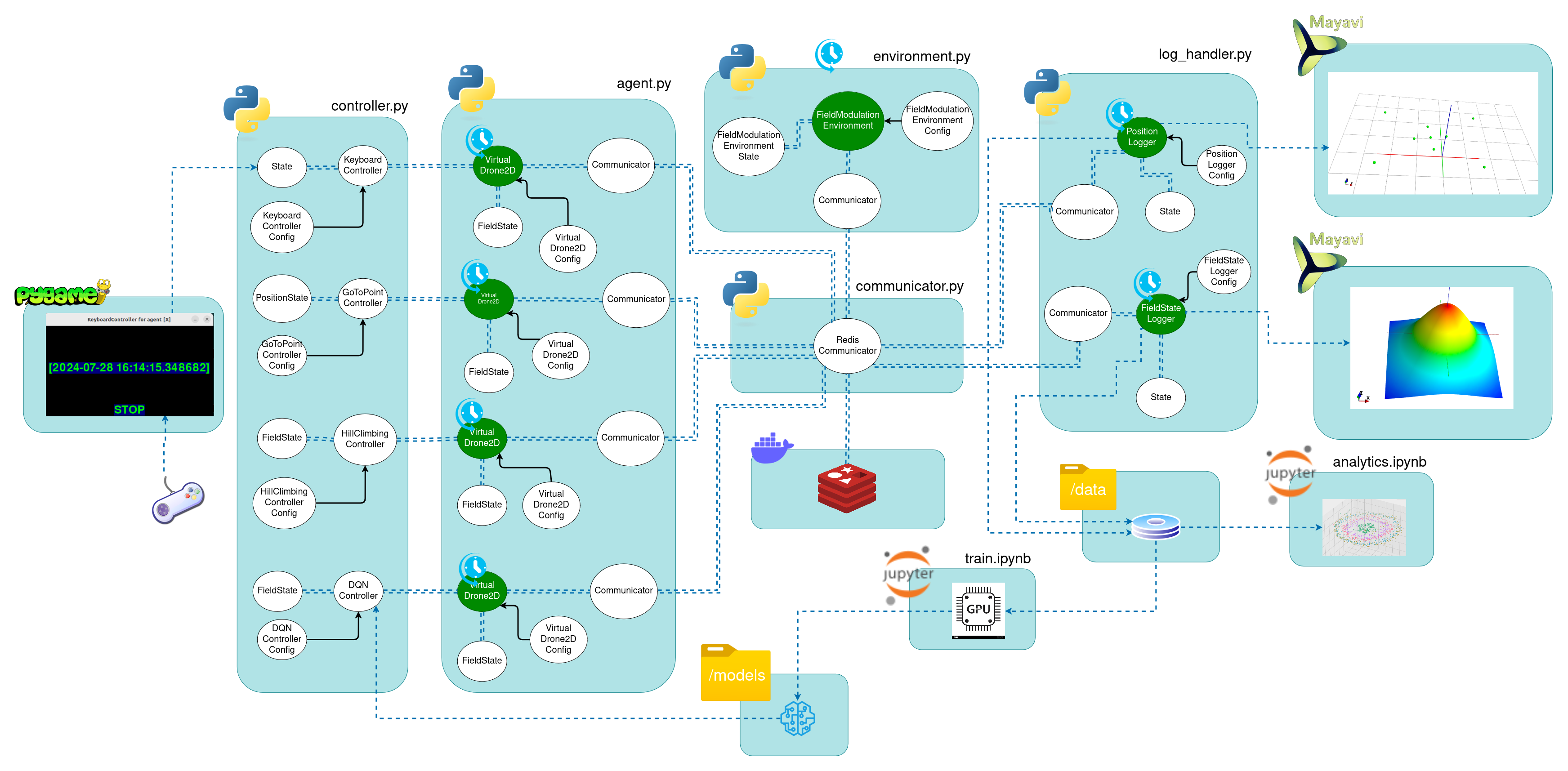}
  \caption{Default implementation}
  \label{fig:default}
\end{figure}

is the most complex implementation done so far and it shows the capabilities of the framework. Best way to traverse Fig.~\ref{fig:default} is to start from the \textbf{Controller}(s) (which solely hold the algorithmic logic of the drones) up to \textbf{LogHandler}(s) (which visualize the information on the network and persist them to disk).

\textbf{KeyboardController} record keyboard strokes withing a \textbf{pygame pop-up} and translate them to 9 discrete actions (8 directions and STOP) for any compatible \textbf{Agent} to use.

\textbf{GoToPointController} is a simple implementation that will match current position with the configured target position in a 2D space and choose to go towards the direction of the maximum delta between X and Y axis in order to reach the target position.

\textbf{HillClimbingController} uses as input data an 84x84 field array (contained in the \textbf{FieldState} object), which will be thoroughly explained in the next section \ref{sec:field-modulation}. Currently it suffice to say the input map is pooled down to a final 3x3 grid by applying a reward function (So far, the `sum` reward function has been proven to be reliable. Future work might consider other choices). Eventually, the cell index with the highest reward from the 3x3 grid is selected, mapped to the one of the 9 discrete actions (8 directions and STOP) and provided to the calling \textbf{Agent}.

\textbf{DQNController} is present in the diagram but at the time of writing it hasn't been implemented due to time constraints. Inspiration for this \textbf{Controller} comes from the success of \textbf{deep Q-network}(s) reported in \cite{mnih2015human} on atari games and UAV path-planning through reinforcement learning in \cite{yan2020towards}. Many choices of this paper such as the 2D space represented by 84x84 map with discrete actions were made in accordance to the previous mentioned works in such a way to facilitate a smooth extension of current work with the desired \textbf{DQNController}. \textbf{KeyboardController}, \textbf{GoToPointController} and, most specifically, the \textbf{HillClimbingController} will serve the purpose of generating situational data for the training process (as depicted in the lower part of Fig.~\ref{fig:default}) in either simulation using the \textbf{VirtualDrone2D} agent or in the physical world through experiments using \textbf{CFDrone2d} (integration with Crazyflie drones explained in section \ref{sec:crazyflie-agent-integration}).

\textbf{VirtualDrone2D} agent starts from an initial position and loops on its configured delay as dictated by the \textbf{VirtualDrone2DConfig}. It broadcast its state on the network and receives the state of the neighbouring agents and the state of the environment. It uses the position information filtered in his configured vicinity limits to construct a representation of its surroundings as a 84x84 map, considering itself in the middle of the map. Shares this information with its underlying \textbf{Controller} for it to eventually choose an action to perform. It maps the action to velocity commands and, under the simple simulated physics of [\( distance = velocity \cdot time \)], computes the new position and awaits for next cycle.  

\textbf{FieldModulationEnvironment} places the space limits (adjustable parameter which for this work was chosen to match the configuration in the VU Amsterdam Laboratory) a list of points of interest in the network for the agents to subscribe. Current implementation allows for the rotation of these points as presented in Fig~\ref{fig:points-of-interest-circle-around-center} and \ref{fig:points-of-interest-circle-spin}.

\textbf{PositionLogger} receives the states of all the registered agents in the network and the space limits from the environment. It uses \textit{mayavi} for real-time visualization of positions in a 3D space and also writes the information to disk on each cycle. The information on disk can be independently read from the disk and plotted as suggested in the \textbf{analytics.ipynb} section. Thereby, \textbf{PositionLogger} is used to provide real-time visualization and persistence of experiment data. 

\textbf{FieldStateLogger} is configured to read, from the network, the state of only one particular compatible \textbf{Agent} that uses \textbf{FieldState}. The 84x84 map contained in the \textbf{FieldState} is visualized and stored to disk. This particular implementation of \textbf{LogHandler} will help, in future variants of this work, to persist the training data for \textbf{DQNController}.

\subsubsection{Crazyflie agent integration}
\label{sec:crazyflie-agent-integration}

\begin{figure}[ht]
  \centering
  \includegraphics[width=1\textwidth]{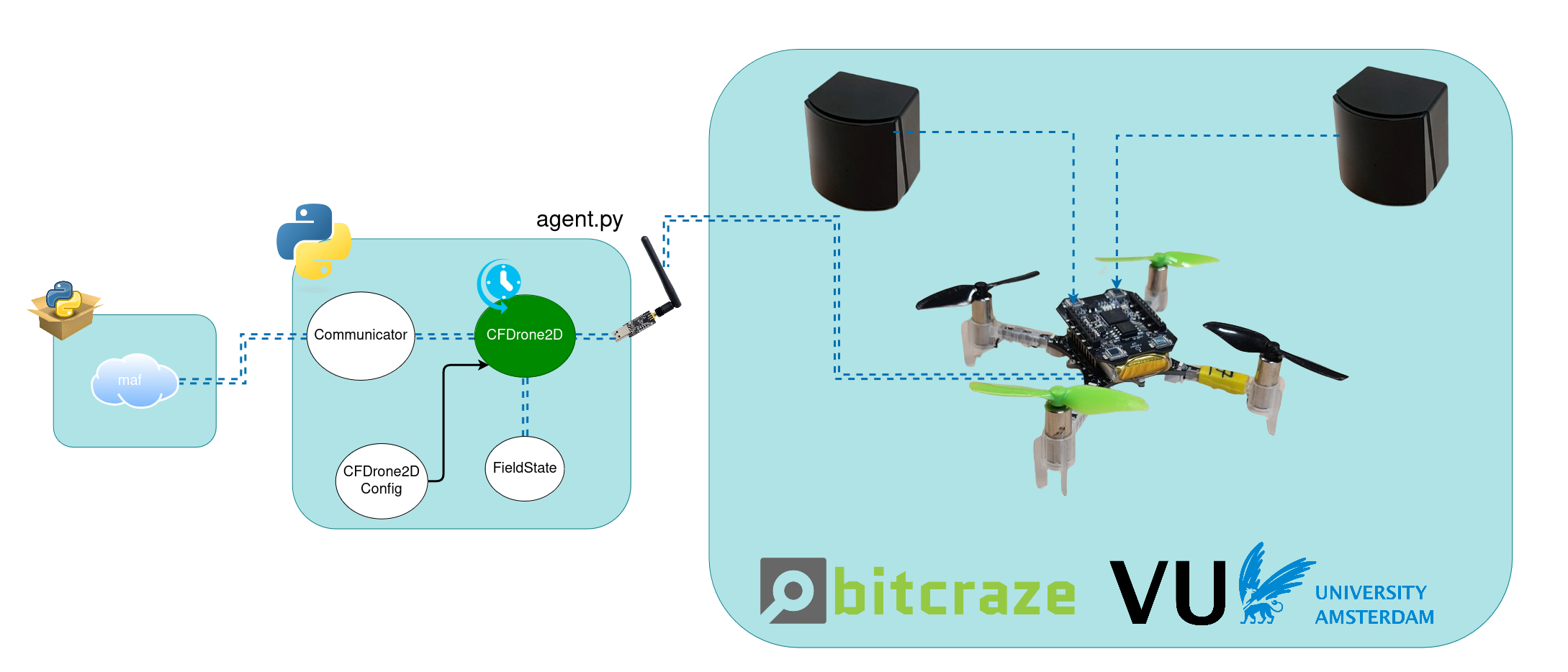}
  \caption{Crazyflie Integration in VU laboratory}
  \label{fig:crazyflie-integration}
\end{figure}

VU Amsterdam laboratory is equipped with \href{https://www.bitcraze.io/products/crazyflie-2-1/}{Crazyflie drones} as shown in Fig.~\ref{fig:crazyflie-integration}.

STM32 microcontroller is used for flight stabilization whilst nRF51 microcontroller is used to communicate information back (receiving velocity commands) and forth (broadcasting position) with \href{https://www.bitcraze.io/products/crazyradio-2-0/}{Crazyradio} commanded by the  \textbf{CFDrone2D} agent implementation in the framework).

VU Amsterdam laboratory has installed the \href{https://www.bitcraze.io/documentation/tutorials/getting-started-with-lighthouse/}{Lighthouse positioning system} having the SteamVR Base stations mounted in the corners of a metal cage thus yielding the space limits used in simulation and real flights. The picture completes with the \href{https://www.bitcraze.io/products/lighthouse-positioning-deck/}{Lighthouse positioning deck} which receives information from the SteamVR Base stations, computes its position and sends it to STM32 using UART protocol. 

\textbf{CFDrone2D} implementation follows same interaction pattern with the rest of the framework just as \textbf{VirtualDrone2D} does, thus they can be used interchangeably. \textbf{CFDrone2D} abstracts away all the necessary adaptation to \href{https://www.bitcraze.io/documentation/repository/}{Bitcraze libraries} to ensure necessary functionalities: \textit{Taking off}, \textit{Receiving position information}, \textit{Sending velocity commands} and \textit{Safe landing}.

\begin{figure}[ht]
  \centering
  \includegraphics[width=1\textwidth]{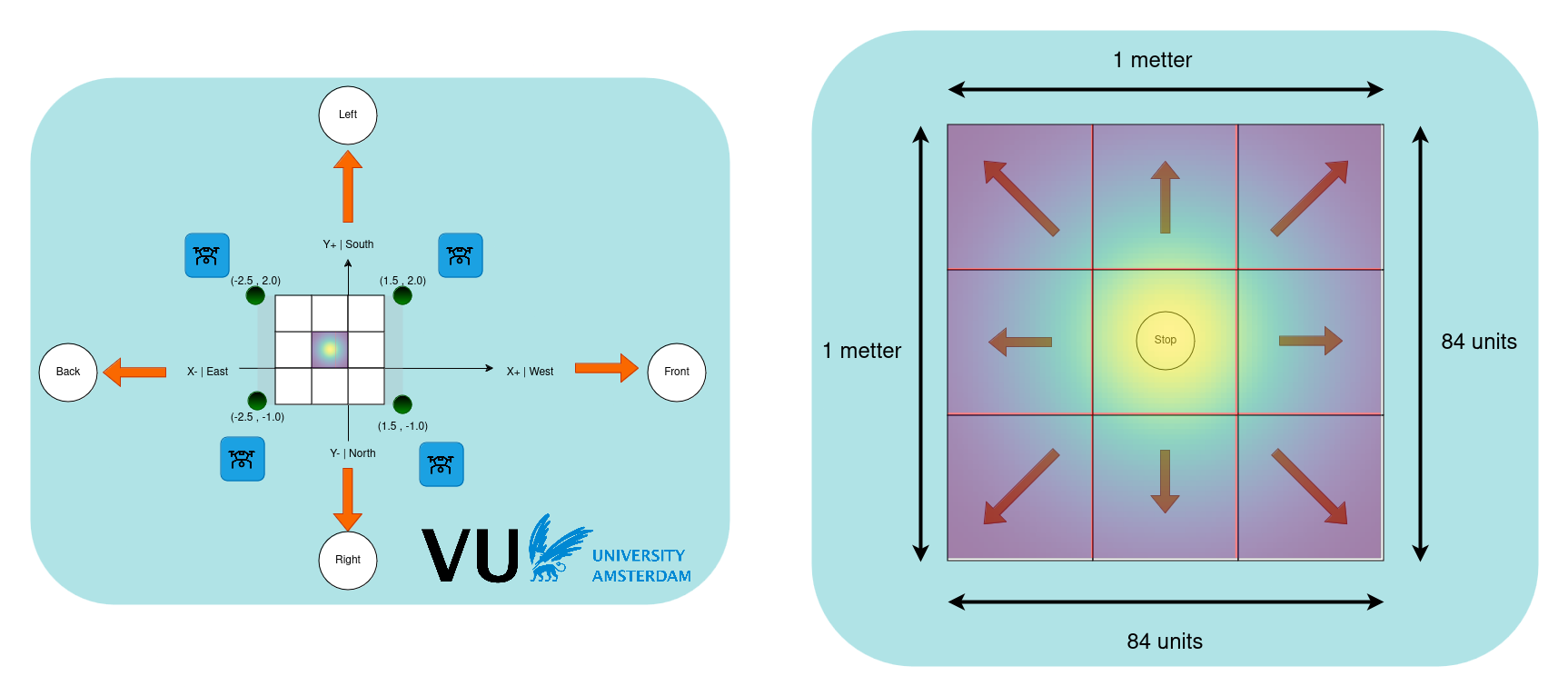}
  \caption{Physical world conventions}
  \label{fig:physical-world-conventions}
\end{figure}

For bridging the sim-to-real gap following conventions described in Fig.~\ref{fig:physical-world-conventions} have been chosen.

Geographic coordinate system  was mapped to a cartesian coordinate system in such a way that:
\begin{itemize}
    \item West direction corresponds to positive X axis and going Front direction;
    \item South direction corresponds to positive Y axis and going Left direction; 
    \item All the other mappings are obvious: e.g. North-East corresponds to negative X and negative Y and going Back-Right direction.
\end{itemize}

Green dots, representing drones, have been placed at the limits of the space as given by the configuration of the laboratory positioning system and same values were used during simulation. 

Right side image in Fig.~\ref{fig:physical-world-conventions} shows how the local 2D vicinity of an agent is being discretized in order to choose between the 9 possible actions.

\subsection{Field Modulation}
\label{sec:field-modulation}

Inspired from the \cite{yan2020towards} situational maps, this paper proposes the real-time construction of such maps with same 84x84 resolution. The main difference implemented in current work is that the maps are centered on the agent in question and they represent its local perception.

Initially, the 84x84 map is filled with values of 0. As current work is concerned, an agent knows the positions of neighbouring agents and the points of interest dictated by the environment, all with respect to a unique global reference frame. The agent will then filter all the points in his local vicinity (0.5 meters on all directions) and project them into its local perception map. The process of applying gaussian transformation centered in this projected points will be called \textit{field modulation} throughout this work.

\subsubsection{Negative field modulation} 
\begin{figure}[ht]
    \centering
    \begin{minipage}{0.2\textwidth}
        \centering
        \includegraphics[width=\linewidth]{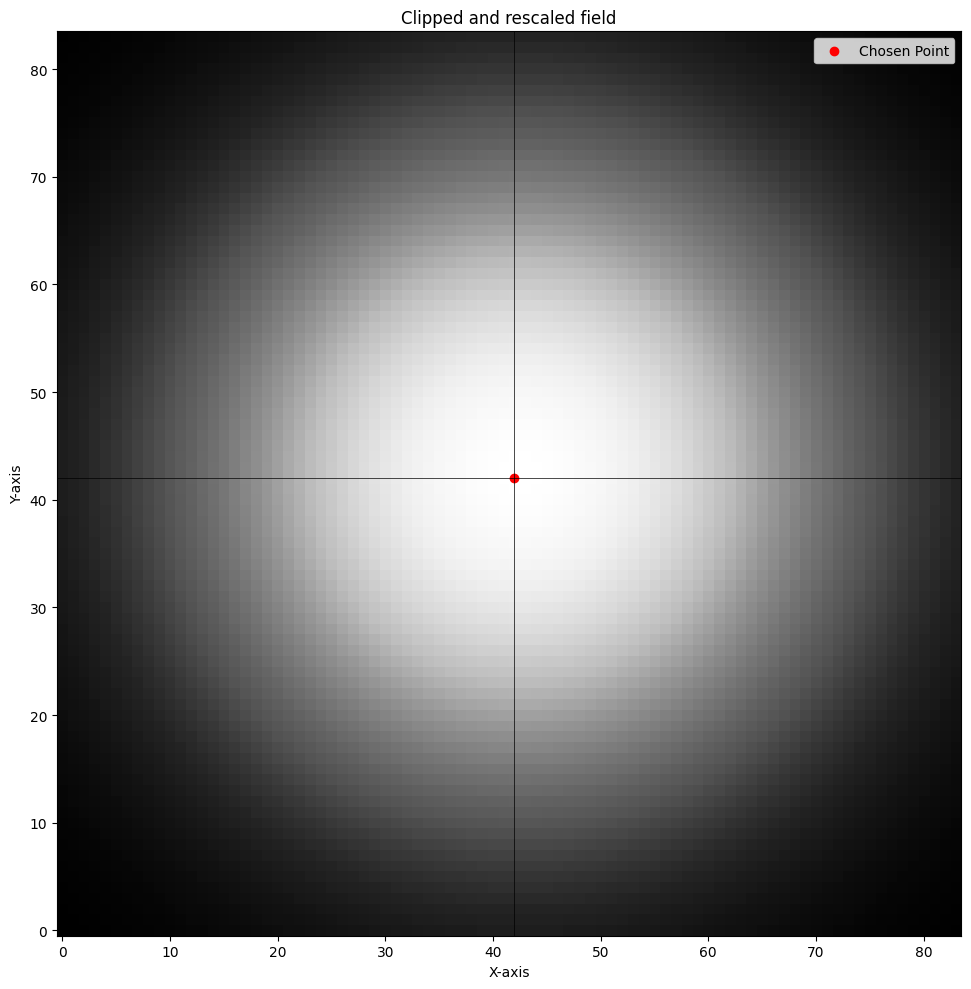}
    \end{minipage}\hfill
    \begin{minipage}{0.3\textwidth}
        \centering
        \includegraphics[width=\linewidth]{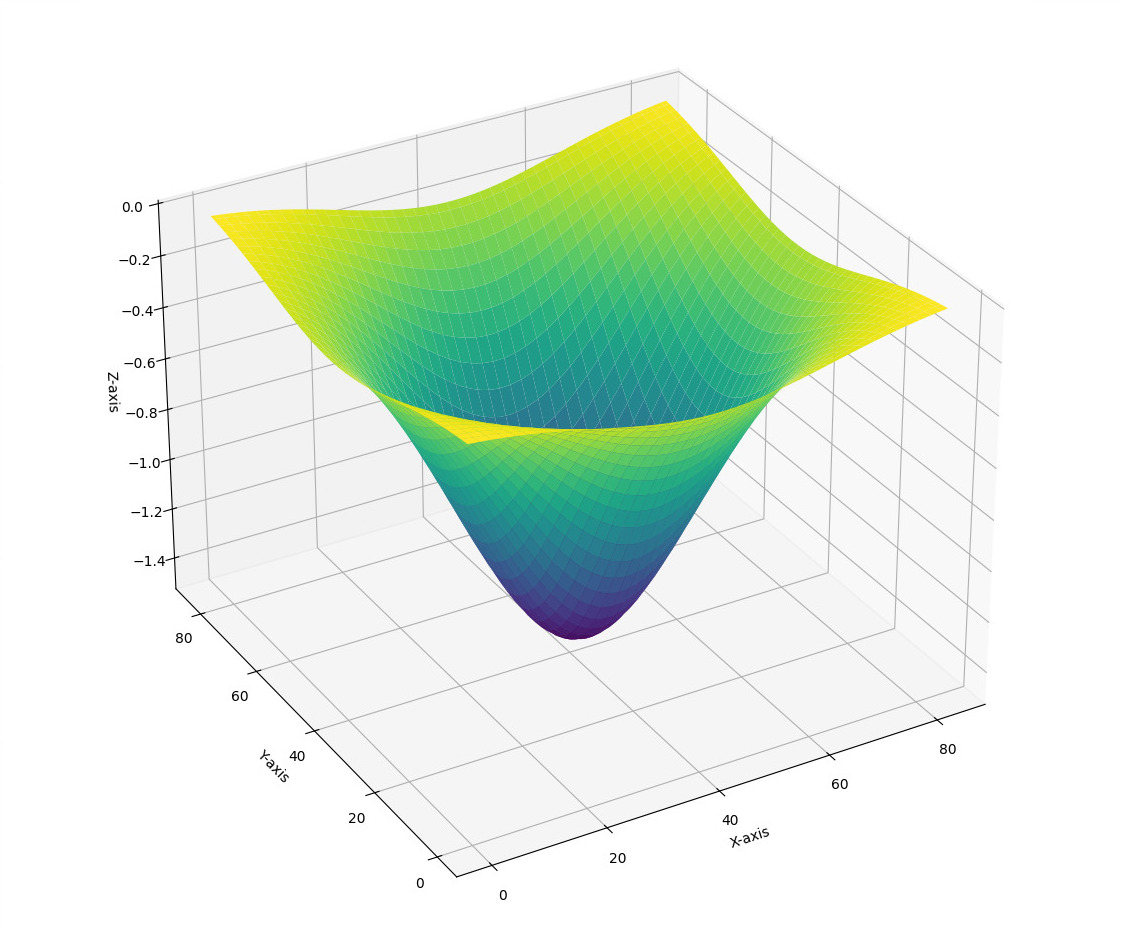}
    \end{minipage}\hfill
    \begin{minipage}{0.45\textwidth}
        \centering
        \includegraphics[width=\linewidth]{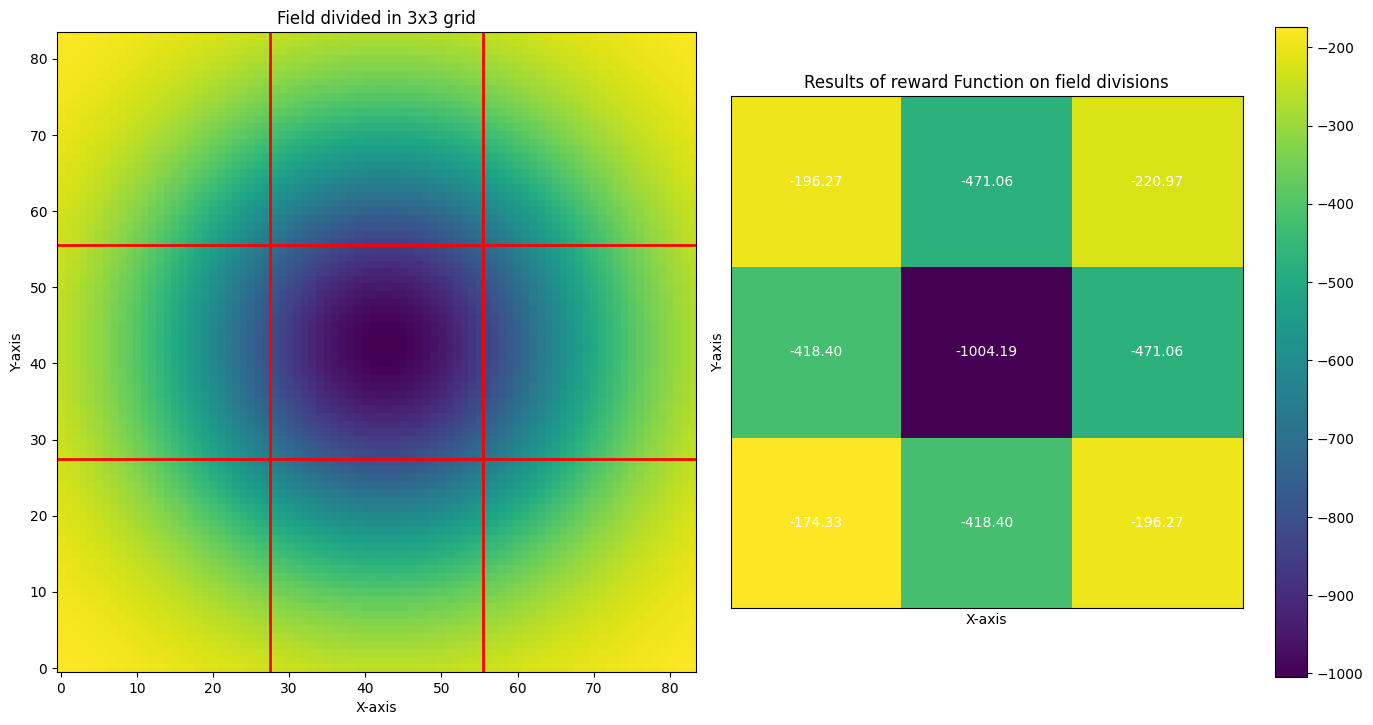}
    \end{minipage}
    \caption{Negative field modulation}
    \label{fig:negative-field-modulation}
\end{figure}

is the process of applying negative gaussians with a maximum depth of 1.5 and a radius of 0.5 meters. Fig.~\ref{fig:negative-field-modulation} describes the case of having a neighbouring agent exactly in the middle of the map. For insights purposes, on the right side it is shown how the \textbf{HillClimbingController} would try to go any other direction in order to escape the negative reward on current position (As will be presented shortly, this is the case of a collision between two agents).

\subsubsection{Positive field modulation} 
\begin{figure}[ht]
    \centering
    \begin{minipage}{0.2\textwidth}
        \centering
        \includegraphics[width=\linewidth]{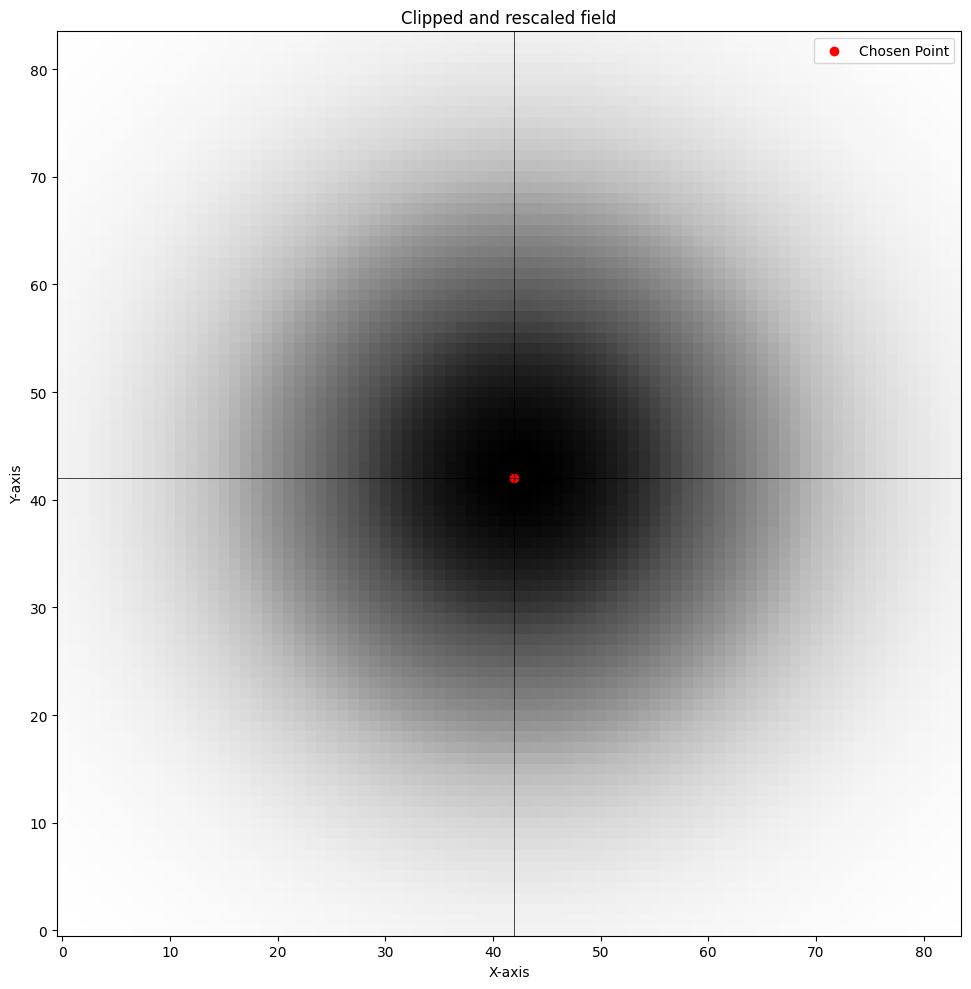}
    \end{minipage}\hfill
    \begin{minipage}{0.3\textwidth}
        \centering
        \includegraphics[width=\linewidth]{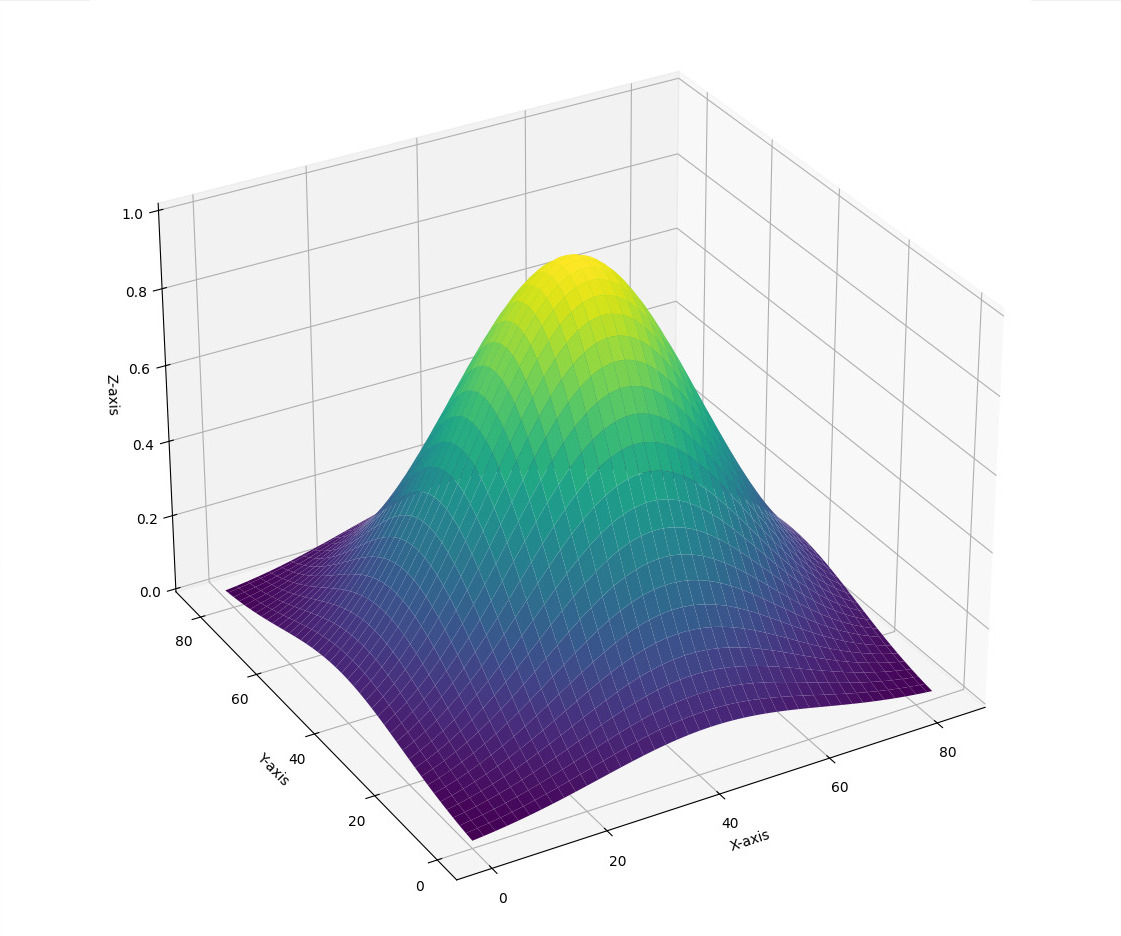}
    \end{minipage}\hfill
    \begin{minipage}{0.45\textwidth}
        \centering
        \includegraphics[width=\linewidth]{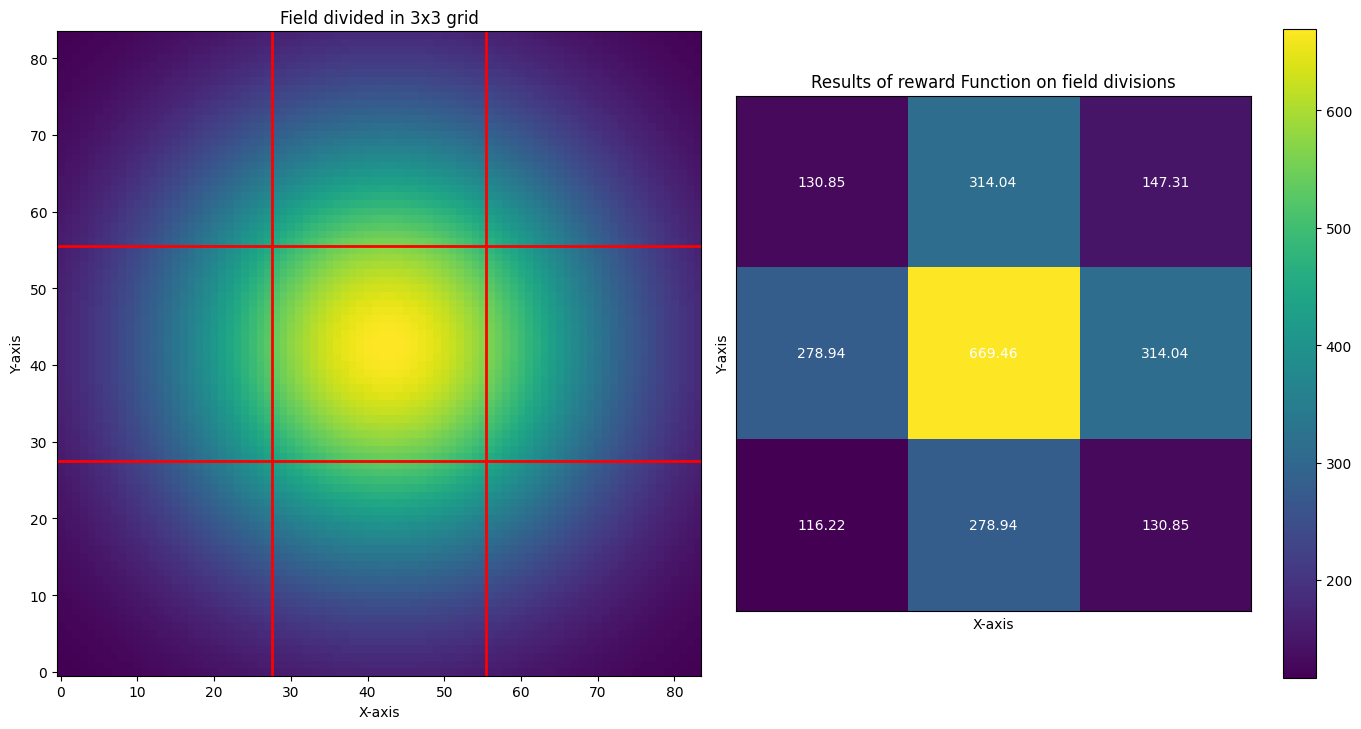}
    \end{minipage}

    \caption{Positive field modulation}
    \label{fig:positive-field-modulation}
\end{figure}

is the process of applying positive gaussians with a maximum height of 1.0 (In terms of absolute values, this is less than the negative 1.5 and was chosen such that avoiding collisions are more important than acquiring high rewards) and a radius of 0.5 meters.
Fig.~\ref{fig:positive-field-modulation} describes this case of having a point of interest (placed by the environment) exactly in the middle of the map. Rightmost image shows that, for example, the \textbf{HillClimbingController} would choose to keep his current position (Stop) in order to maximize the reward.

\subsubsection{Collision avoidance} 
\begin{figure}[ht]
    \centering
    \begin{minipage}{0.48\textwidth}
        \centering
        \includegraphics[width=\linewidth]{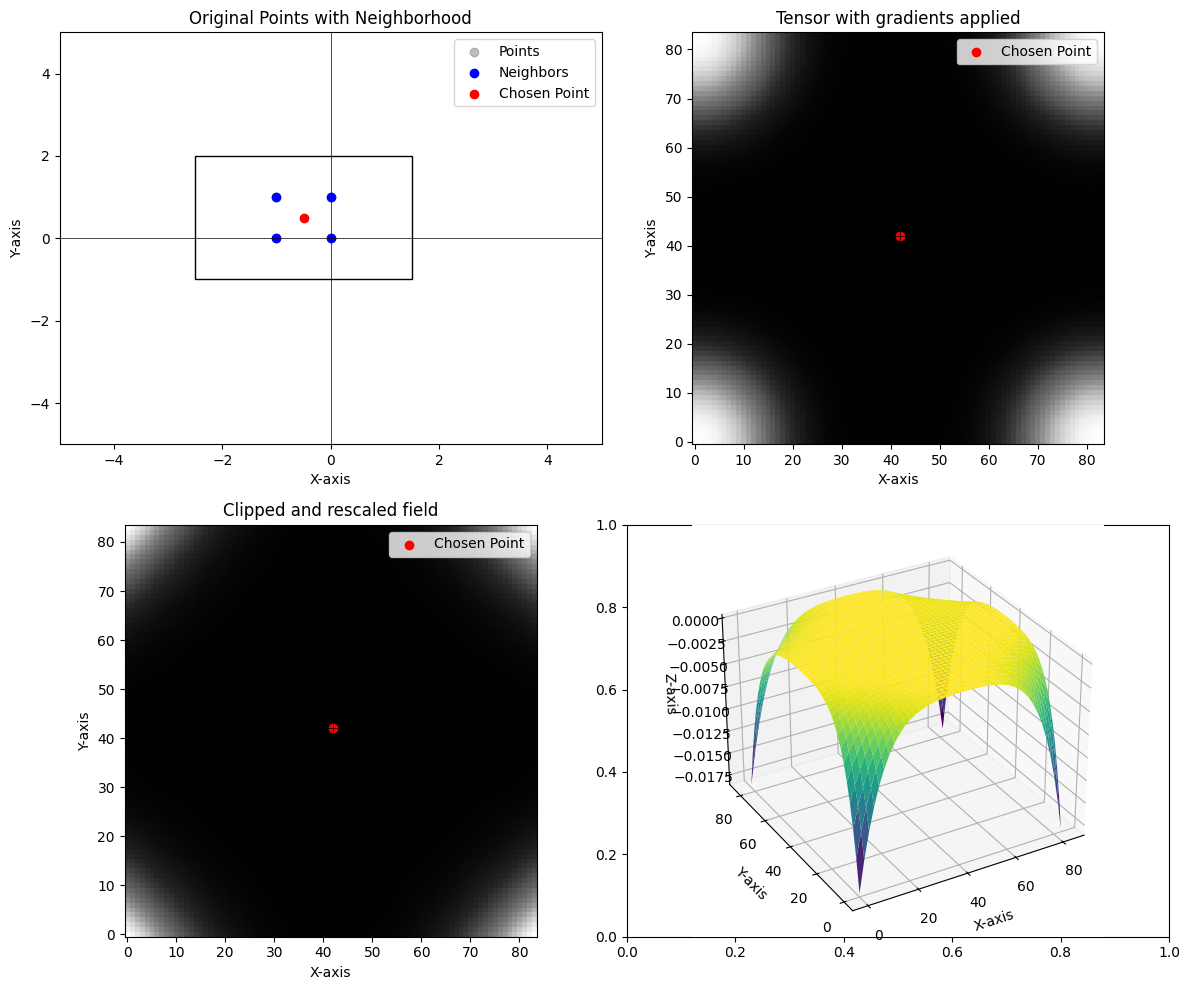}
    \end{minipage}
    \hfill
    \begin{minipage}{0.48\textwidth}
        \centering
        \includegraphics[width=\linewidth]{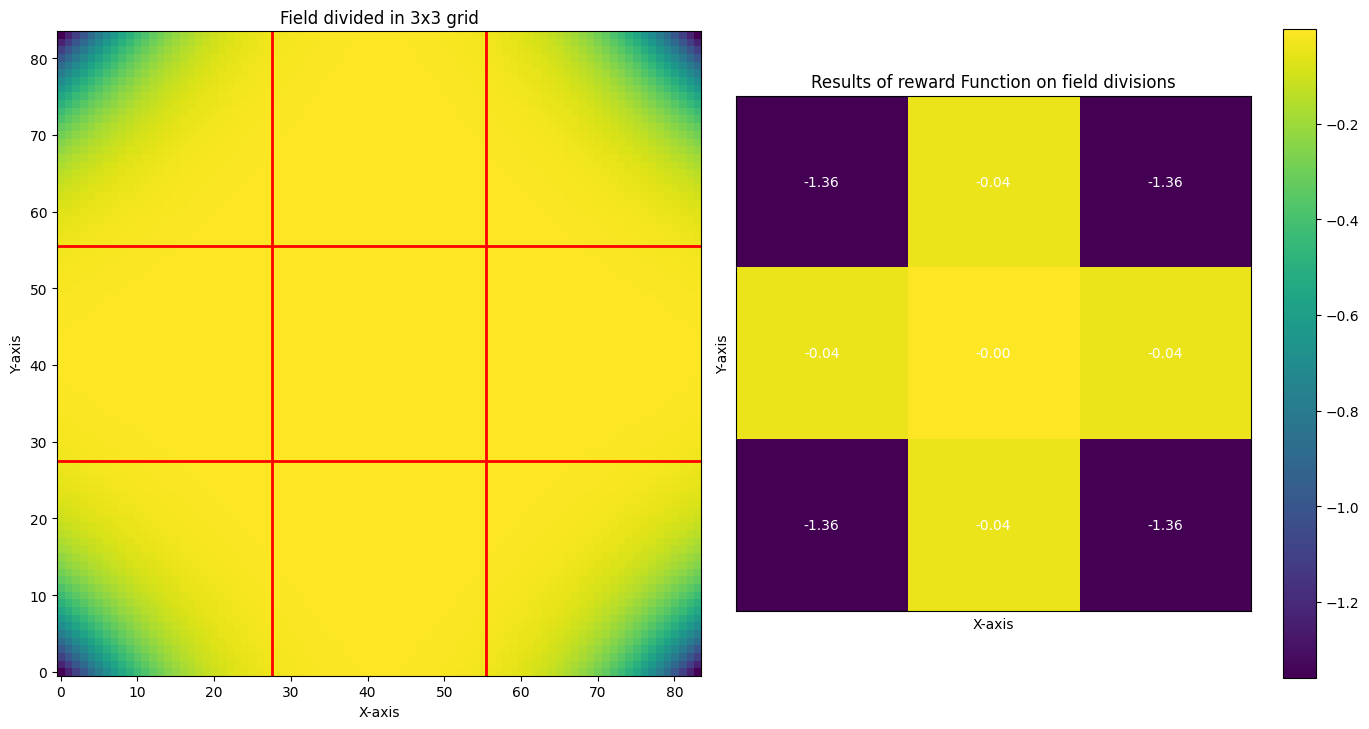}
    \end{minipage}
    
    \caption{Collision avoidance}
    \label{fig:collision-avoidance}
\end{figure}

is the first hypothesis to be tested. In order to better describe it, the scenario in Fig.~\ref{fig:collision-avoidance} has been simulated.

In the first image, the agent in question is the red dot and has the 4 agents in his local vicinity. The neighbouring agents are projected into the 84x84 map and negative field modulation is performed. The map is clipped and rescaled based on an adjustable \textit{Clip size factor} (2.0 in this case) so that the agent can perceive the effect of modulations applied outside his final limits of perception. Eventually, the visualization and grid divisions of \textbf{HillClimbingController} suggest that the agent will avoid going towards the corners where the collision with neighbours might occur.

\subsubsection{Gradient following} 
\begin{figure}[ht]
    \centering
    \begin{minipage}{0.48\textwidth}
        \centering
        \includegraphics[width=\linewidth]{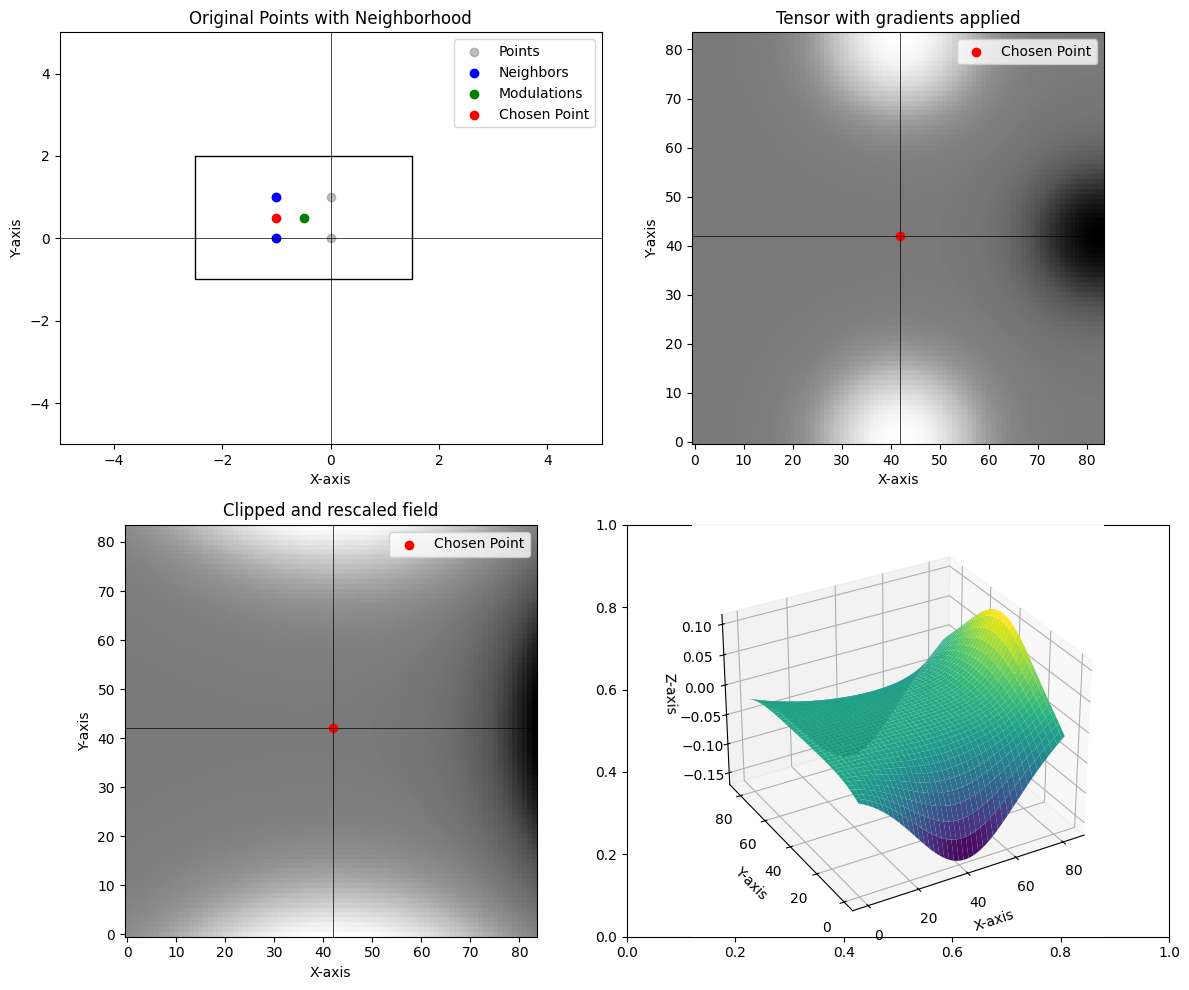}
    \end{minipage}
    \hfill
    \begin{minipage}{0.48\textwidth}
        \centering
        \includegraphics[width=\linewidth]{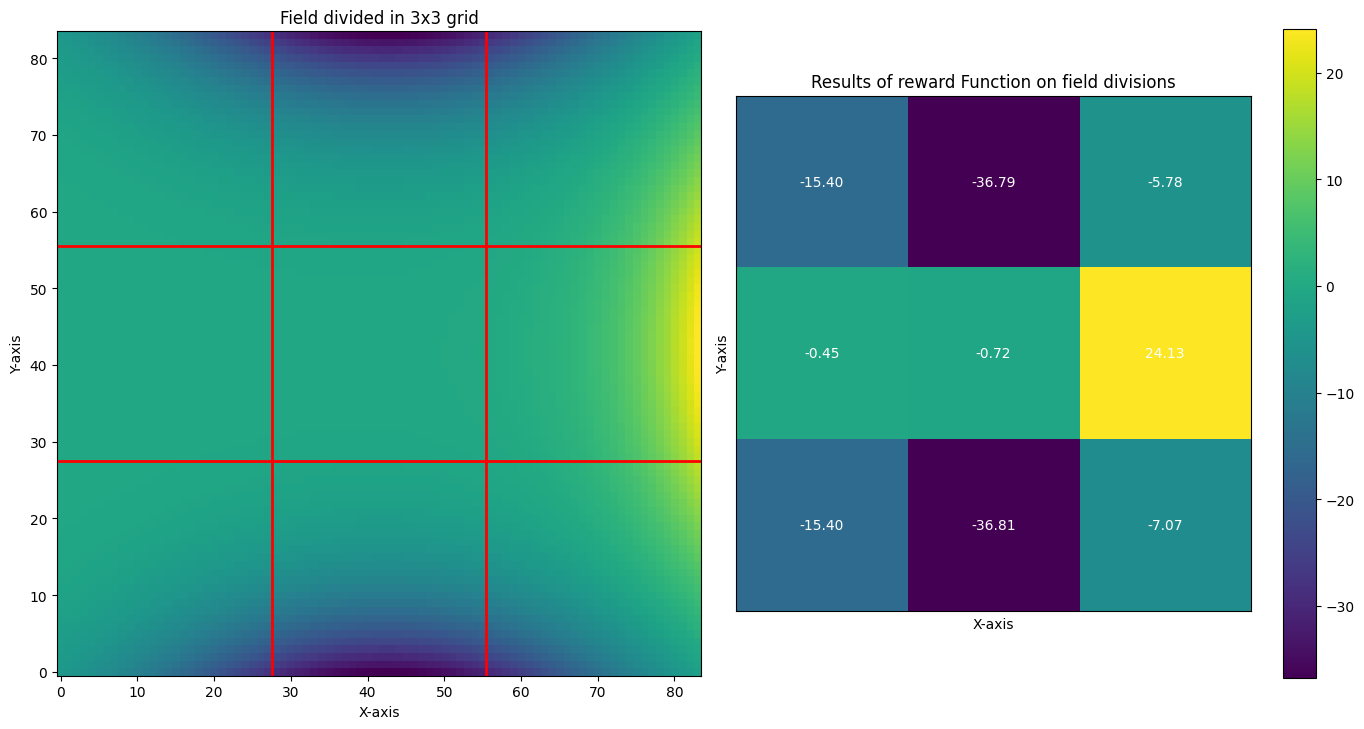}
    \end{minipage}
    \caption{Gradient following}
    \label{fig:gradient-following}
\end{figure}

is the second hypothesis which will be tested in simulation and physical world experiments. The scenario in Fig.~\ref{fig:gradient-following} presents 4 agents positioned in the corners of a square and the main agent represented as the red dot is placed in the middle of one side, thereby perceiving only 2 agents as his neighbours.

In the middle of the square, the point of interest (green dot), placed by the environment, is close enough for the the red agent to perceive and to consequently apply a positive field modulation there. Although not present in this simulation, when a negative and positive field modulations affect same region in the map, the effect is cumulative. Again, the \textbf{HillClimbingController} insights suggest that the red agent will choose to go Front direction (positive X axis), following the high reward and avoid going left or right towards neighbouring agents.

\subsubsection{Boundary limits} 
\begin{figure}[ht]
    \centering
    \begin{minipage}{0.48\textwidth}
        \centering
        \includegraphics[width=\linewidth]{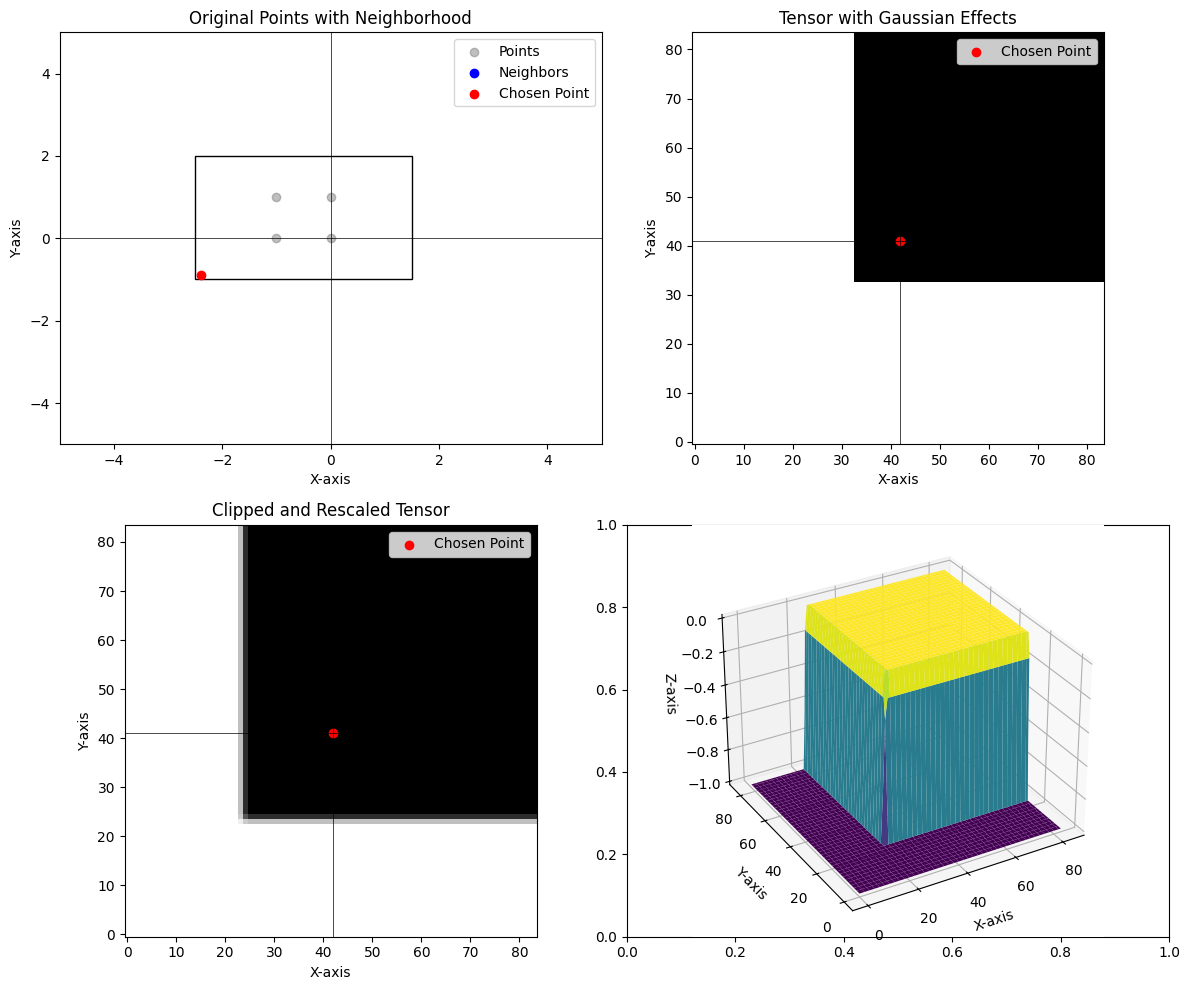}
    \end{minipage}
    \hfill
    \begin{minipage}{0.48\textwidth}
        \centering
        \includegraphics[width=\linewidth]{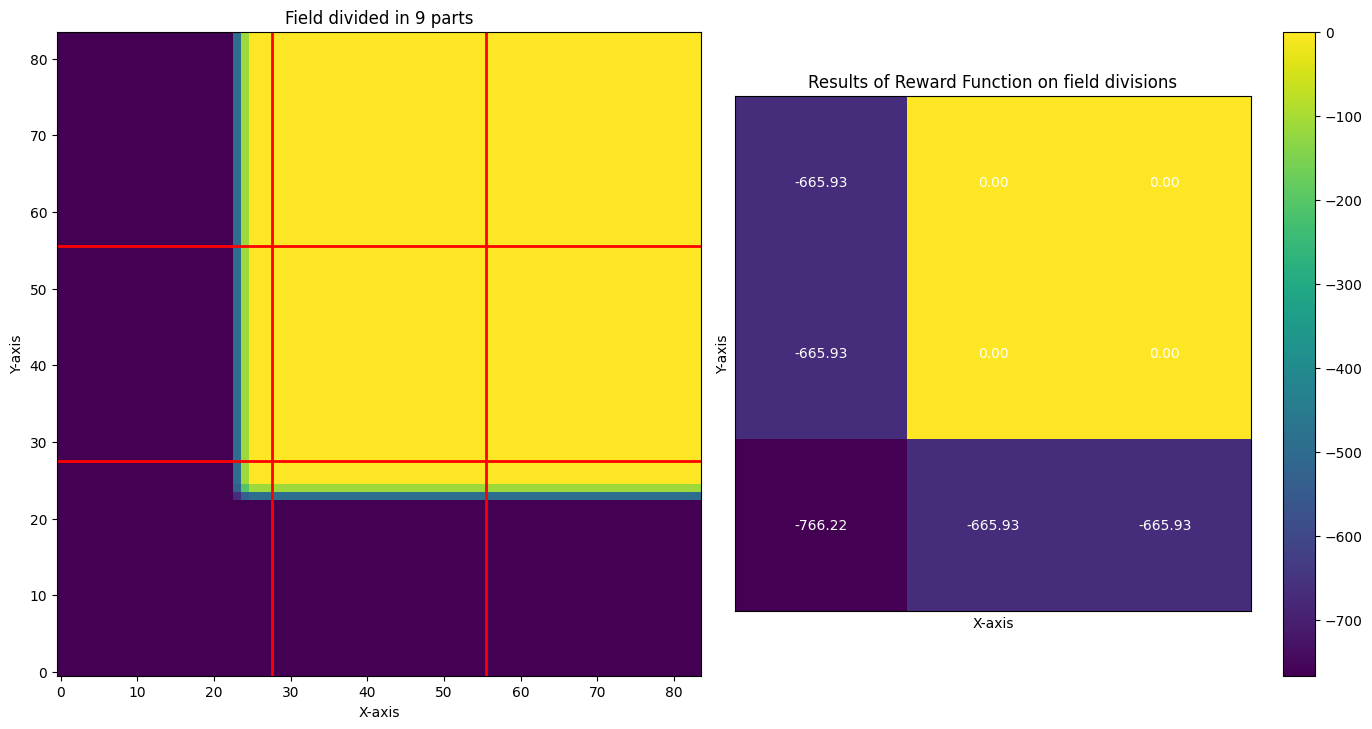}
    \end{minipage}

    \caption{Boundary limits}
    \label{fig:boundary-limits}
\end{figure}

is the last hypothesis that the agents will keep themselves out of the boundaries imposed by the environment. Fig.~\ref{fig:boundary-limits} presents such a simulation.

The red agent will always try to project the space limits into its local perception map and, when there is an overlap, it will apply a patch with a value of -1.0. Insights from \textbf{HillClimbingController} suggest that the agent should choose the neutral rewards of 0.0 in spite of the negative rewards perceived outside the space limits. Such a mechanism is needed to ensure safe experimentation in the physical world.

\section{Results}

Final experiments will be presented in depth in sections \ref{sec:circle-around-center-section} and \ref{sec:circle-spin-section}. All the information required to replicate the experiments is found in their corresponding configuration files (See source code).

In both sections parameters have been tuned in simulation with the use of \textbf{VirtualDrone2D} agents and further deployed in the physical world through the use of \textbf{CFDrone2D} agents. Current work did not asses the quality of the sim-to-real gap with numerical means but can only state that, on a visual inspection, the behaviour of agents in the physical world followed closely the one in the simulation. 

\subsection{Circle around center}
\label{sec:circle-around-center-section}
\begin{center}
\href{https://alexandru-dochian.github.io/multi_agent_framework/pages/circle_around_center.html}{Click to watch experiment clip}
\end{center}

\begin{figure}[ht]
    \centering
    \begin{minipage}{0.3\textwidth}
        \centering
        \includegraphics[width=\linewidth]{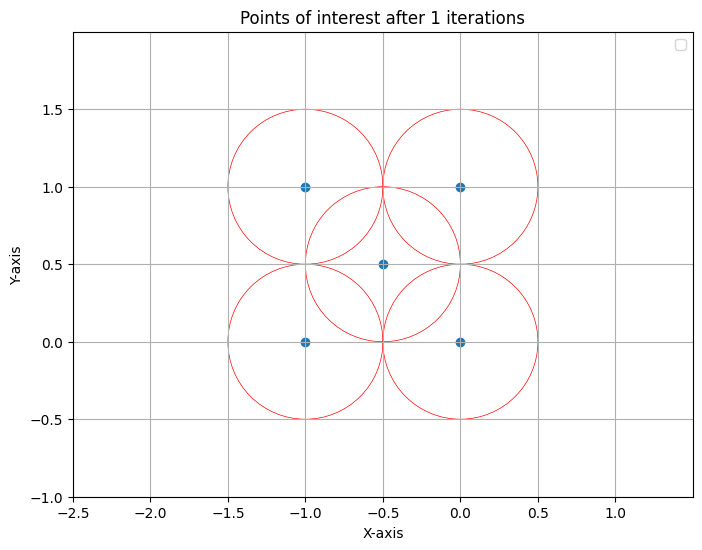}
    \end{minipage}\hfill
    \begin{minipage}{0.3\textwidth}
        \centering
        \includegraphics[width=\linewidth]{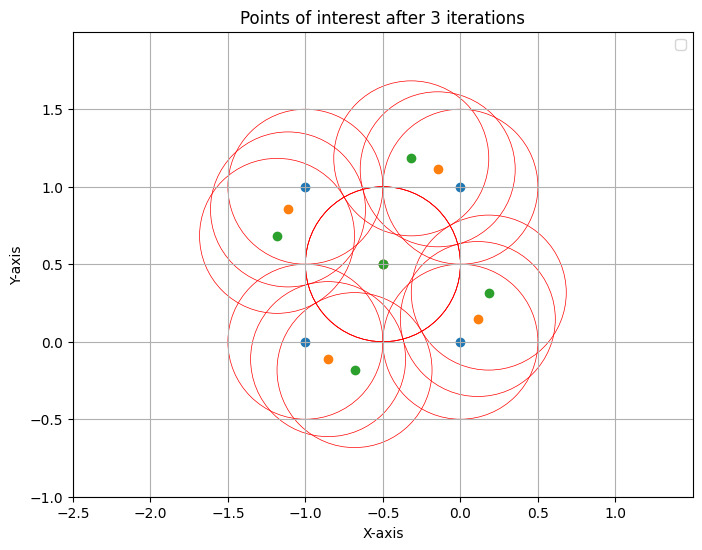}
    \end{minipage}\hfill
    \begin{minipage}{0.3\textwidth}
        \centering
        \includegraphics[width=\linewidth]{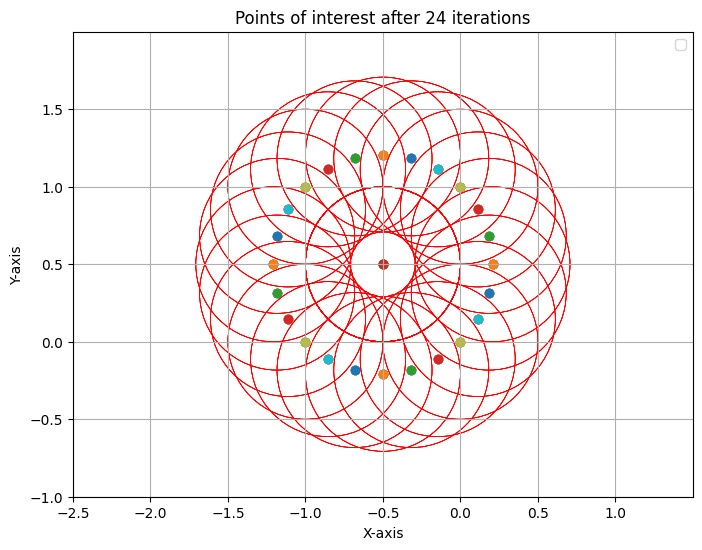}
    \end{minipage}
    \caption{Points of interest placed by the environment during circle around center experiment}
    \label{fig:points-of-interest-circle-around-center}
    \centering
    \begin{minipage}{0.49\textwidth}
        \centering
        \includegraphics[width=\linewidth]{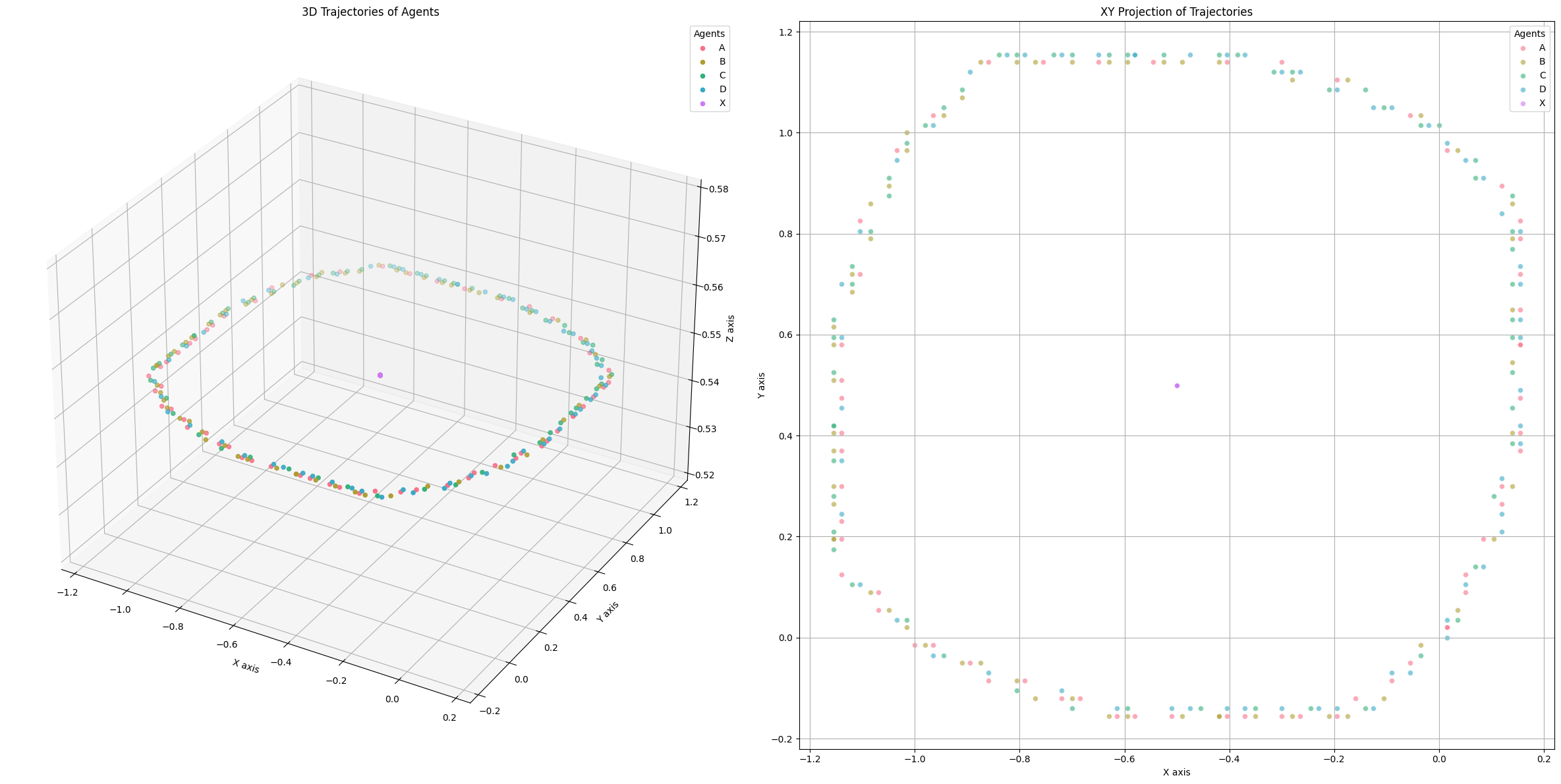}
        \caption{Trajectories of agents during \textbf{Circle around center} experiment -- in simulation}
        \label{fig:circle_around_center}
    \end{minipage}\hfill
    \begin{minipage}{0.49\textwidth}
        \centering
        \includegraphics[width=\linewidth]{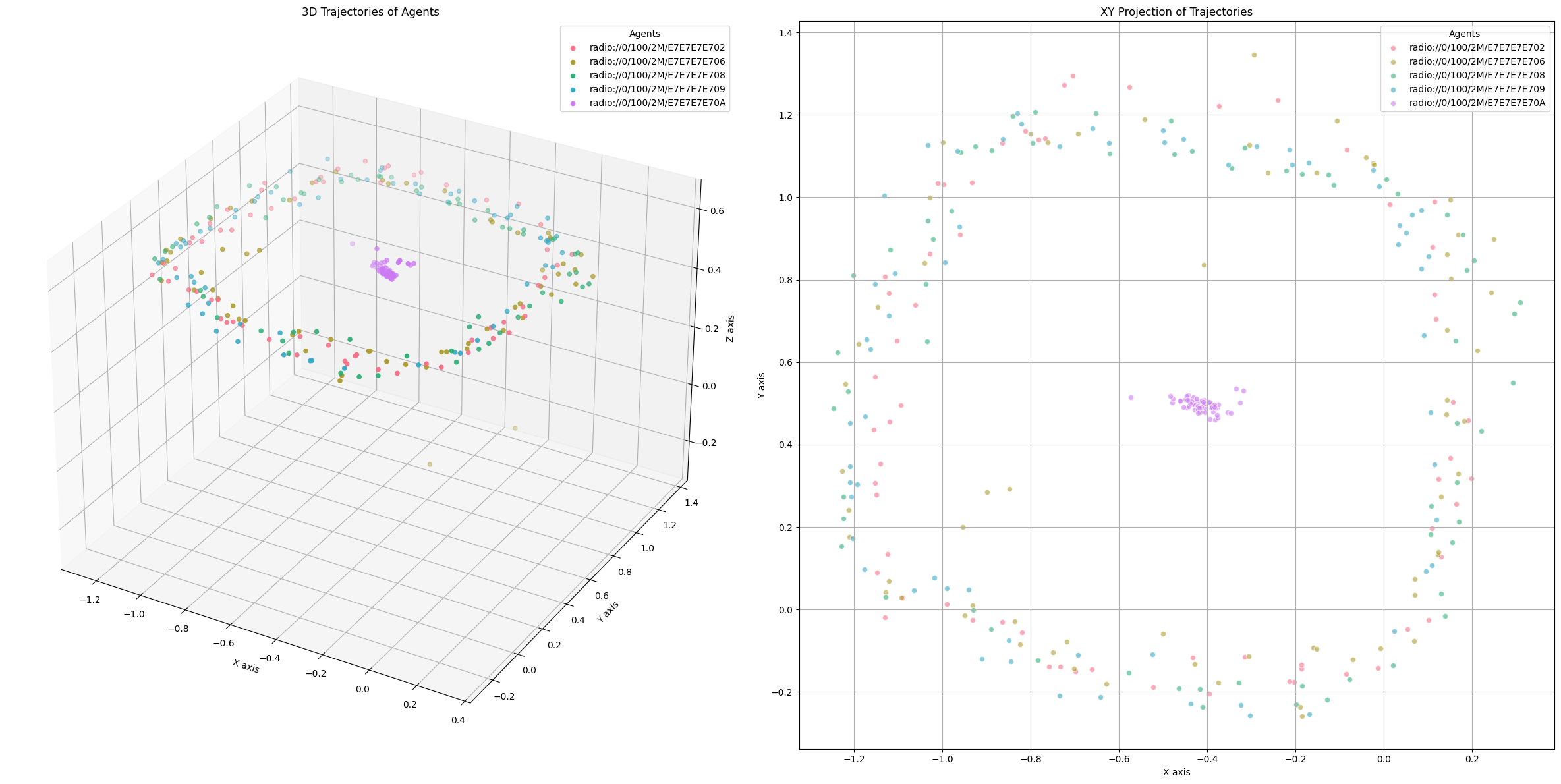}
        \caption{Trajectories of agents during \textbf{Circle around center} experiment -- in physical world}
        \label{fig:circle_around_center_vu}
    \end{minipage}
    \centering
\end{figure}

Circle around center uses 5 agents starting from random positions. Points of interest were rotated by the environment during the experiment and sent over the network for the agents to receive as described in Fig.~\ref{fig:points-of-interest-circle-around-center}.

Experiment results from Fig.~\ref{fig:circle_around_center} were generated by executing the \\ \textbf{circle\_around\_center.json} configuration file and made use of \textbf{VirtualDrone2D} agents having following identifiers: \textbf{A}, \textbf{B}, \textbf{C}, \textbf{D} and \textbf{X}.

Experiment results from Fig.~\ref{fig:circle_around_center_vu} were generated by executing the \\ \textbf{circle\_around\_center\_vu.json} configuration file and made use of \textbf{CFDrone2D} agents having following unique identifiers: \textbf{O2}, \textbf{06}, \textbf{08}, \textbf{09}, \textbf{0A}.

Both the simulated and real experiment used the same parameters tuned in simulation and presented in table \ref{table:env_agent_params}

\begin{table}[htbp]
\centering
\begin{minipage}{0.48\textwidth}
\centering
\begin{tabular}{|l|c|}
\hline
\multicolumn{2}{|c|}{\textbf{Environment Parameters}} \\ \hline
\textbf{Delay}           & 100 ms          \\ \hline
\textbf{Limit X (min)}   & -2.5         \\ \hline
\textbf{Limit X (max)}   & 1.5          \\ \hline
\textbf{Limit Y (min)}   & -1.0         \\ \hline
\textbf{Limit Y (max)}   & 2.0          \\ \hline
\textbf{Modulations}     & [-1, 1], [0, 1], [0, 0] \\
                & [-1, 0], [-0.5, 0.5] \\ \hline 
\textbf{Rotation center} & [-0.5, 0.5] \\ \hline
\textbf{Theta}           & 2.5 $^\circ$ \\ \hline
\end{tabular}
\end{minipage}%
\hspace{0.01\textwidth}
\begin{minipage}{0.48\textwidth}
\centering
\begin{tabular}{|l|c|}
\hline
\multicolumn{2}{|c|}{\textbf{Agent Parameters}} \\ \hline
\textbf{Delay}               & 100 ms            \\ \hline
\textbf{Vicinity Limit}      & {[}0.5 meters, 0.5 meters{]} \\ \hline
\textbf{Field Size}          & 84 x 84        \\ \hline
\textbf{Clip Size Factor}    & 2.0            \\ \hline
\textbf{Reward Function}     & sum            \\ \hline
\textbf{Default Height}      & 0.55 meters    \\ \hline
\textbf{Velocity}            & 0.35 meters    \\ \hline
\end{tabular}
\end{minipage}
\vspace{0.5cm} 
\caption{Environment and Agent parameters in circle around center experiment}
\label{table:env_agent_params}
\end{table}

\vspace{0.5cm} 

\subsection{Circle spin}
\label{sec:circle-spin-section}
\begin{center}
\href{https://alexandru-dochian.github.io/multi_agent_framework/pages/circle_spin.html}{Click to watch experiment clip}
\end{center}

Circle spin experiment also uses 5 agents starting from random positions.

\begin{figure}[ht]
    \centering
    \begin{minipage}{0.3\textwidth}
        \centering
        \includegraphics[width=\linewidth]{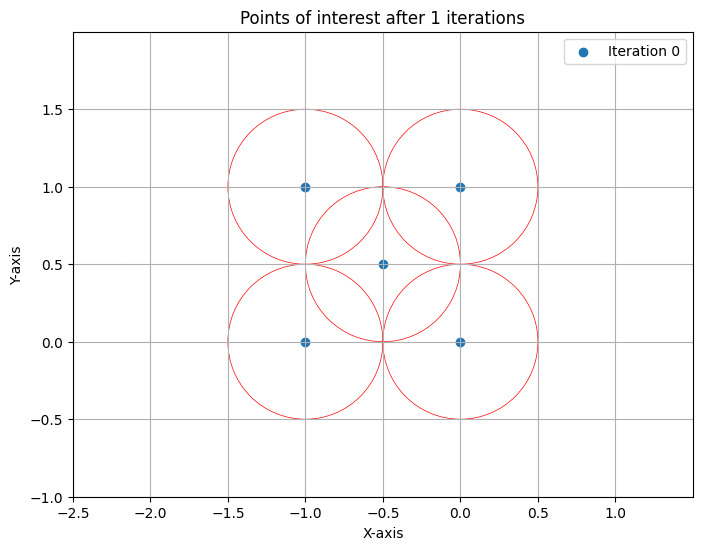}
    \end{minipage}\hfill
    \begin{minipage}{0.3\textwidth}
        \centering
        \includegraphics[width=\linewidth]{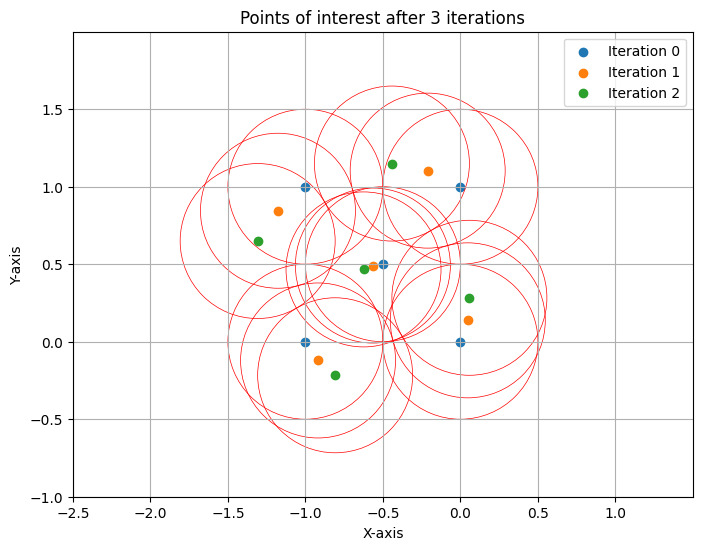}
    \end{minipage}\hfill
    \begin{minipage}{0.3\textwidth}
        \centering
        \includegraphics[width=\linewidth]{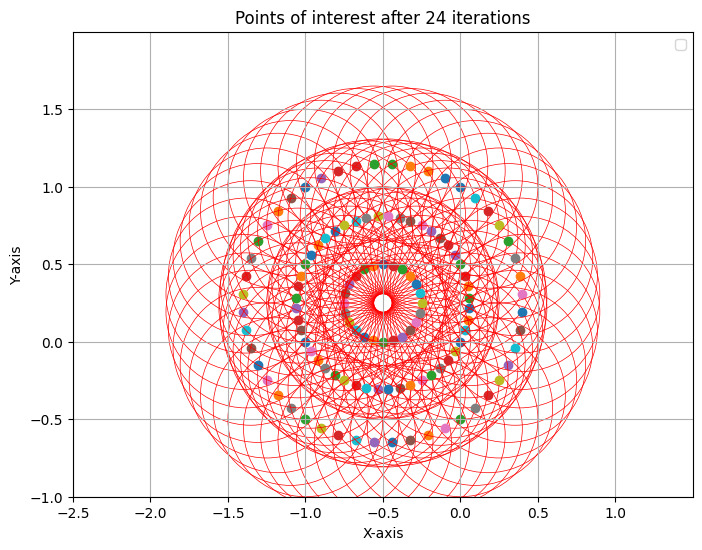}
    \end{minipage}
    \caption{Points of interest placed by the environment during circle spin experiment}
    \label{fig:points-of-interest-circle-spin}
    \centering
    \begin{minipage}{0.49\textwidth}
        \centering
        \includegraphics[width=\linewidth]{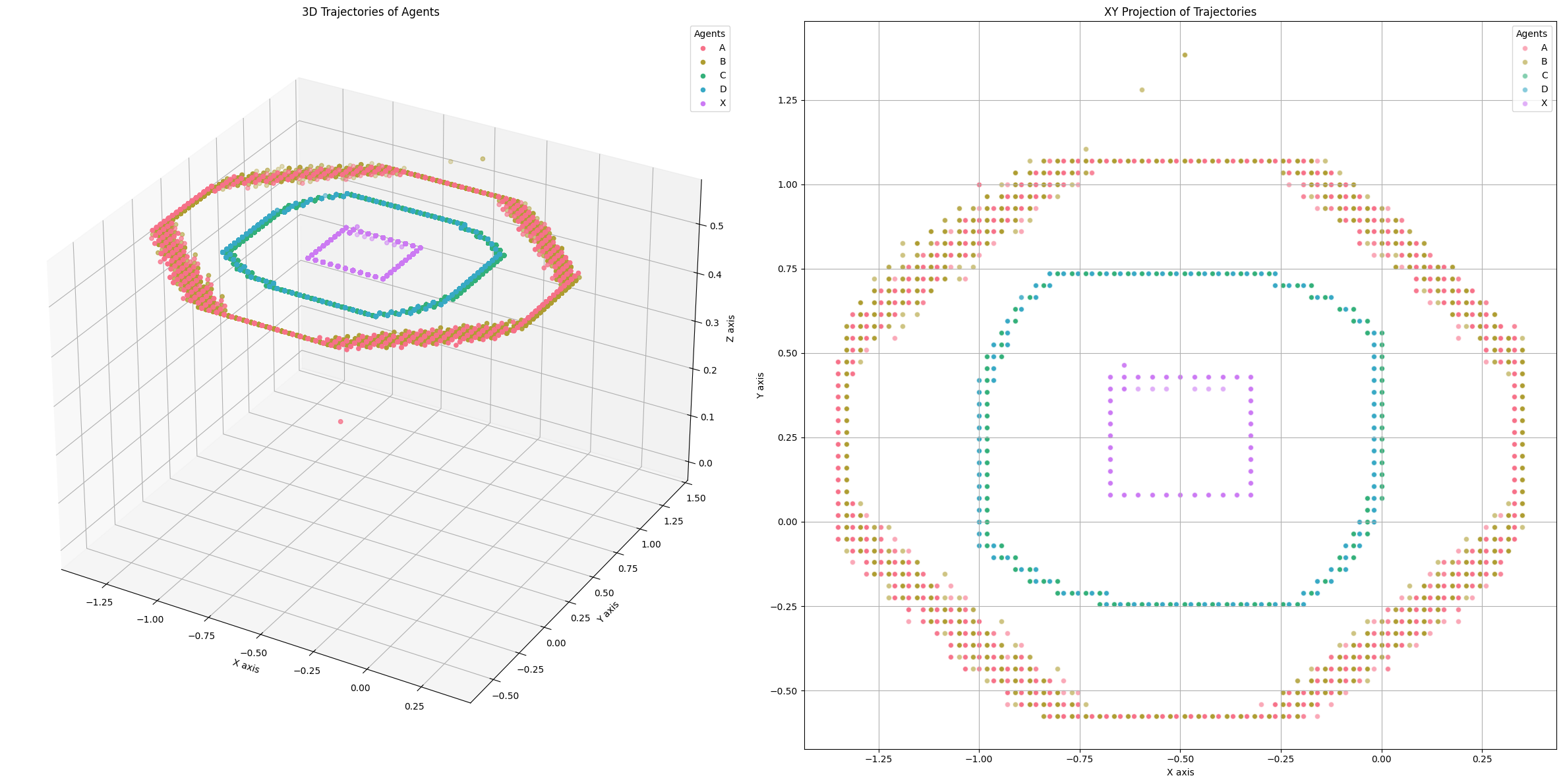}
        \caption{Trajectories of agents during \textbf{Circle spin} experiment -- in simulation}
        \label{fig:circle_spin}
    \end{minipage}\hfill
    \begin{minipage}{0.49\textwidth}
        \centering
        \includegraphics[width=\linewidth]{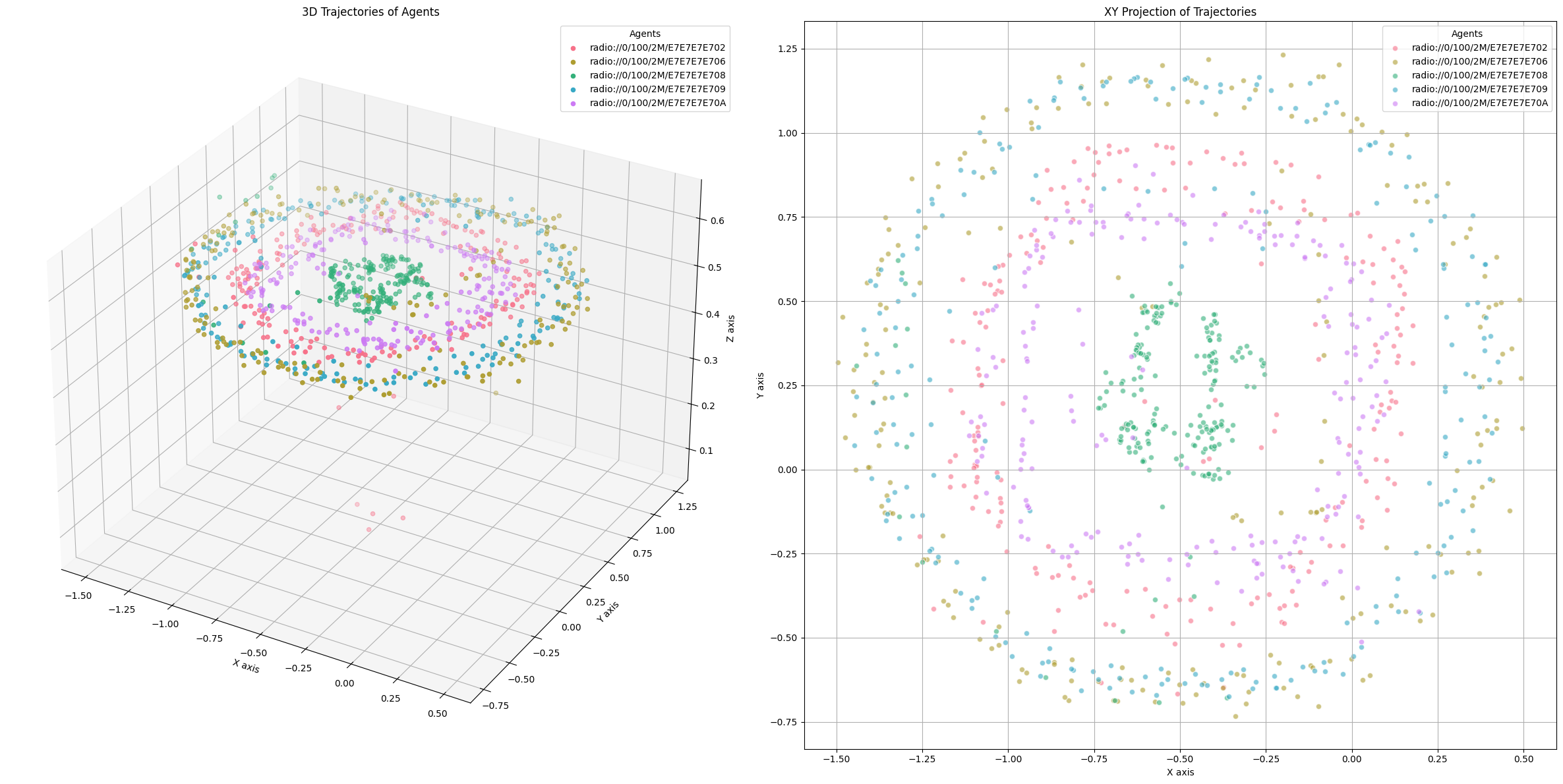}
        \caption{Trajectories of agents during \textbf{Circle spin} experiment -- in physical world}
        \label{fig:circle_spin_vu}
    \end{minipage}
    \centering
\end{figure}

This one is almost identical to the aforementioned \textbf{Circle around center} experiment with \textbf{the only difference} of having a different \textbf{Rotation center} in the environment configuration, namely \textbf{[-0.5, 0.25]} instead of the \textbf{[-0.5, 0.5]}. This slight modification accounts for an interesting emerging pattern as presented in Fig.~\ref{fig:points-of-interest-circle-spin}.

Experiment results from Fig.~\ref{fig:circle_spin} were generated by executing the \\ \textbf{circle\_spin.json} configuration file and made use of \textbf{VirtualDrone2D} agents having following identifiers: \textbf{A}, \textbf{B}, \textbf{C}, \textbf{D} and \textbf{X}.

Experiment results from Fig.~\ref{fig:circle_spin_vu} were generated by executing the \\ \textbf{circle\_spin\_vu.json} configuration file and made use of \textbf{CFDrone2D} agents having following unique identifiers: \textbf{O2}, \textbf{06}, \textbf{08}, \textbf{09}, \textbf{0A}.

\section{Conclusion}

In this work, a multi-agent framework—or, more technically, a runtime environment with decentralized execution of computer processes and network communication capabilities—has been developed. It has been shown that it supports a wide range of applications from a simple Hello World application to UAV swarms of crazyflie drones in VU Amsterdam laboratory. Versatility of the software allows for hybrid experimentation with multiple \textit{controllers} such as \textbf{KeyboardController}, \textbf{GoToPointController} and \textbf{HillClimbingController} being used interchangeably by \textbf{VirtualDrone2D} and \textbf{CFDrone2D} \textit{agents}. Also, \textit{environment} implementations such as \textbf{FieldModulationEnvironment} or \textit{log handler} implementations such as \textbf{PositionLogger} or \textbf{FieldLogger} can be wired into compatible experimental setups to enrich them with extra information or features (e.g. real-time visualization). 

The field modulation theory hypothesis of \textbf{Collision avoidance}, \textbf{Gradient following} and \textbf{Boundary limits} were proven to be sound in a short range of experiments. Further evaluative studies are to be conducted to ascertain the limits of this theory.

By embedding all necessary execution information into a \textbf{json configuration file}, the framework ensures that experiments can be reused and replicated in both simulation and physical environments at the expense of pressing a button.

\section{Limitation and Future work}

Current work settled on some constraints in order to facilitate the speedy development of the end-to-end pipeline from theory to safe flight of UAV swarm in the physical world.

\textbf{First limitation} is the 2D environment constraint. Extension to swarming 3D environment would introduces additional challenges. For example, the current 84x84 map that is pooled down to 3x3 grid with 9 discrete actions would possibly be extended to a 84x84x84 map being pooled down to 3x3x3 grid with 27 discrete actions.

The \textbf{second limitation} is the discrete action space, which could potentially be replaced with polar coordinates. This extension might be useful for achieving dynamic velocities, thereby addressing the \textbf{third limitation}: the constant velocity of agents that is present in all current implementations.

As previously mentioned, the \textbf{DQNController} is not implemented at the time of writing. Arguably, future extensions should prioritize this implementation as many decisions in this work were specifically taken to support the implementation of such controller. 

\subsection{Zooming in}
The initial plan for this work was actually what is presented in Fig.~\ref{fig:zoom-in}:

\begin{figure}[ht]
    \centering
    \includegraphics[width=0.9\linewidth]{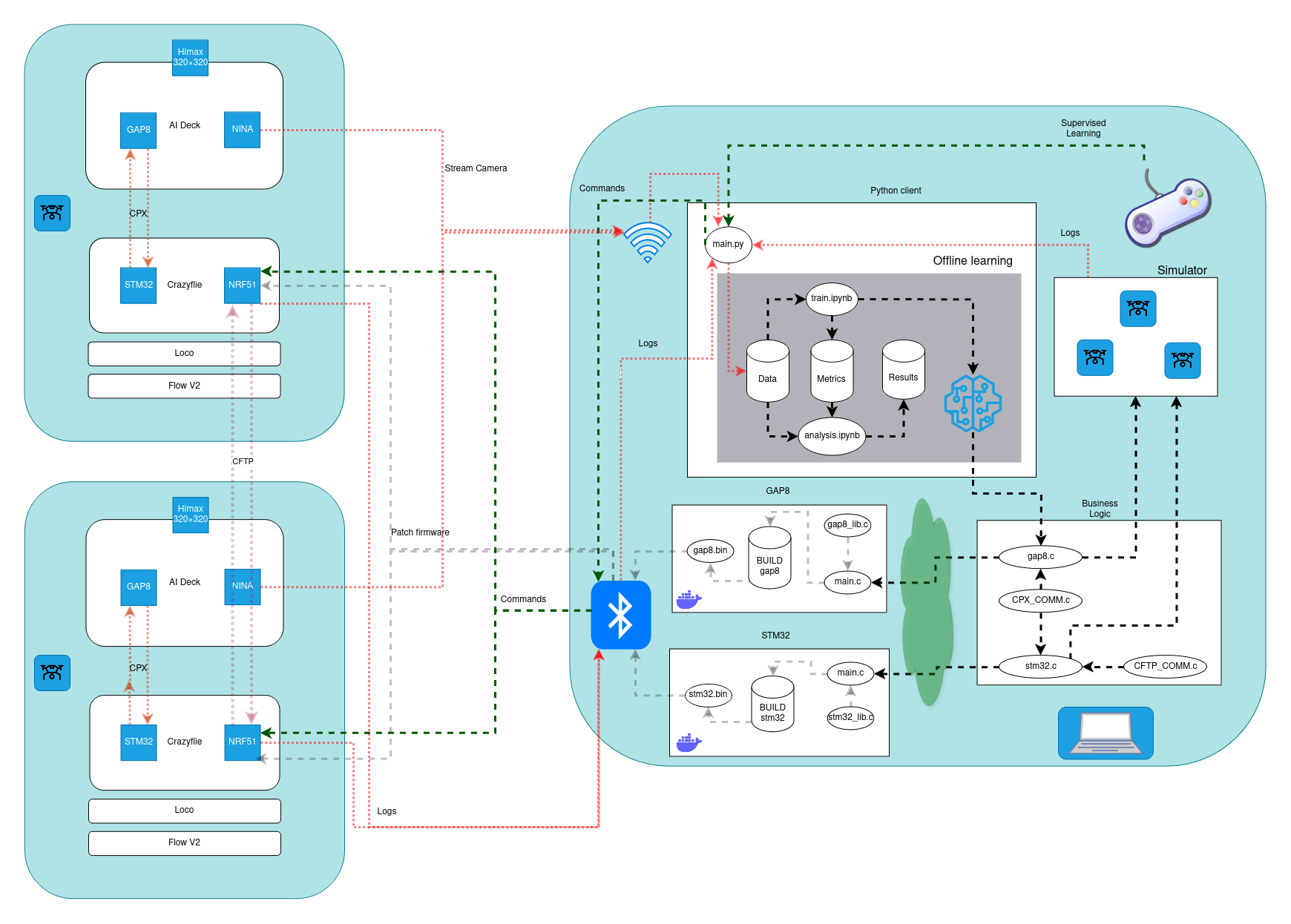}
    \caption{Architecture for onboard control of Crazyflie drone through AI deck}
    \label{fig:zoom-in}
\end{figure}

\begin{itemize}
    \item \href{https://www.bitcraze.io/products/ai-deck/}{AI deck} is responsible for deciding the next action to perform. The GAP8 RISC-V processor is highly optimized for matrix multiplication and can access the information the Himax camera on a fast lane or from the Crazyflie deck via \href{https://www.bitcraze.io/documentation/repository/crazyflie-firmware/master/functional-areas/cpx/}{CPX};
    \item \href{https://www.bitcraze.io/products/crazyflie-2-1/}{Crazyflie deck} is responsible for flight stabilization and performing the actions received from the AI deck;
    \item Depending on the application , the \href{https://www.bitcraze.io/products/loco-positioning-deck/}{Loco} and \href{https://www.bitcraze.io/products/flow-deck-v2/}{Flow V2} decks would provide the position to the the Crazyflie deck (In fact loco positioning system is considerably more inaccurate then the lighthouse positioning system that has been used throughout this work. It is recalled in this section only for documentation purposes);
\end{itemize}

In this scenario the computer responsibilities would be:
\begin{itemize}
    \item Building and flashing the corresponding firmware for GAP8 processor and STM32 microcontroller;
    \item Offline training of neural networks for applications requiring it;
    \item Logging runtime information as position or camera;
\end{itemize}

\subsection{Zooming out}
\begin{figure}[ht]
    \centering
    \includegraphics[width=0.9\linewidth]{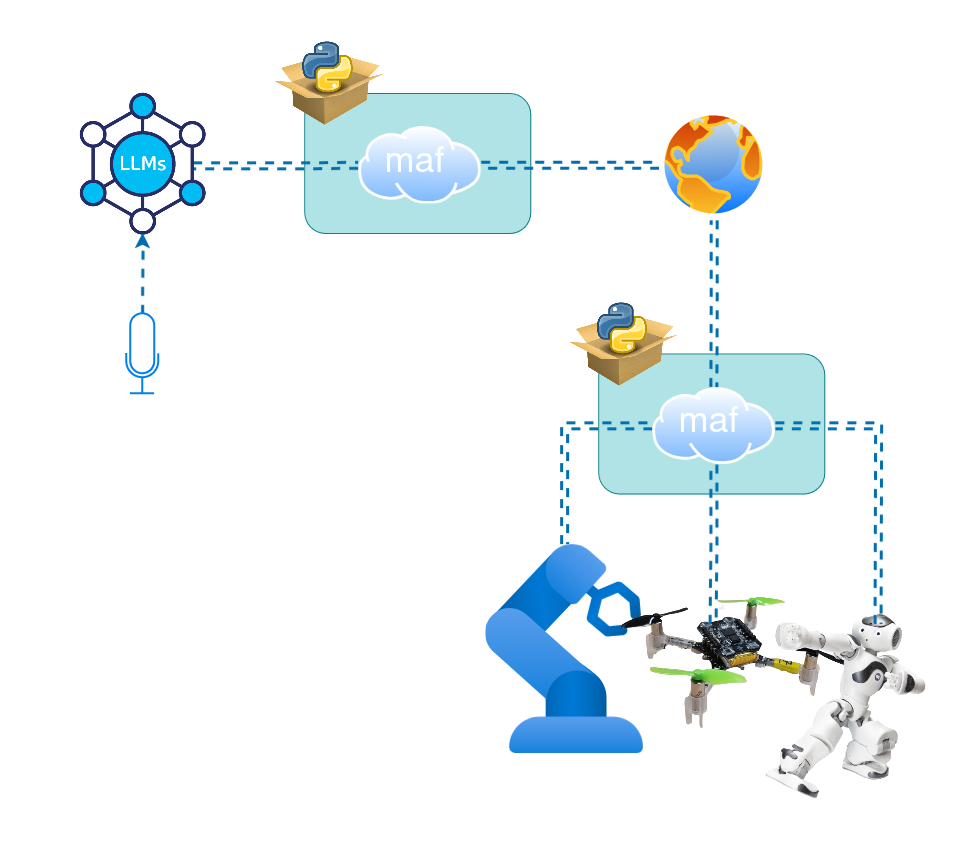}
    \caption{Scenario of a cloud-centric extension of current work}
    \label{fig:zoom-out}
\end{figure}

One sensible extension of the system would be to enhance the cloud support. Fig.~\ref{fig:zoom-out} presents the scenario of a cluster of \textbf{multi agent framework} instances that exchange information over the network.

Potentially, a \textbf{LLMAgent} implementation on one end might receive information, process it and propagate it through the network to interact with agents like \textbf{RoboticArmAgent}, \textbf{NaoRobotAgent} and last but not least \textbf{CFDrone3D}.

\clearpage
\bibliographystyle{splncs04}
\bibliography{references}

\end{document}